\documentclass[aoas]{imsart}

\usepackage{xcolor}
\usepackage[OT1]{fontenc}
\usepackage{amsthm,amsmath}
\usepackage{xr}
\usepackage[figuresright]{rotating}
\usepackage{booktabs,multirow,array}
\usepackage{mathrsfs, bm}
\usepackage{algorithm}
\usepackage{algorithmic}
\usepackage[utf8]{inputenc}
\usepackage[english]{babel}
\usepackage{float}
\usepackage[english]{babel}
\usepackage[utf8]{inputenc}
\usepackage[english]{babel}
\usepackage{natbib}

\startlocaldefs
\newtheorem{propdef}{Definition}
\newtheorem{prop}{Proposition}
\newtheorem{lemma0}{Lemma}

\def\bm{\boldsymbol}

\def\L{\mathcal{L}}

\def\bY{\bm Y}

\DeclareMathOperator*{\argmin}{\arg\!\min}

\theoremstyle{plain} 
\endlocaldefs

\makeatletter
\def\mathcolor#1#{\@mathcolor{#1}}
\def\@mathcolor#1#2#3{%
  \protect\leavevmode
  \begingroup
    \color#1{#2}#3%
  \endgroup
}
\makeatother

\begin{document}

\begin{frontmatter}
\title{LOCUS: A regularized blind source separation method with low-rank structure for investigating brain connectivity}
\runtitle{LOCUS: Low-rank connectivity decomposition with uniform sparsity}

\begin{aug}
\author[A]{\fnms{Yikai} 
\snm{Wang}\ead[label=e1,mark]{ywan566@emory.edu}}
\and
\author[A]{\fnms{Ying} \snm{Guo}\ead[label=e2,mark]{yguo2@emory.edu}}

\address[A]{Department of Biostatistics and Bioinformatics, Emory University, Atlanta, GA 30322\\ \printead{e1,e2}}
\end{aug}

\begin{abstract}
Network-oriented research has been increasingly popular in many scientific areas. In neuroscience research, imaging-based network connectivity measures have become the key for understanding brain organizations, potentially serving as individual neural fingerprints.  There are major challenges in analyzing connectivity matrices including the high dimensionality of brain networks, unknown latent sources underlying the observed connectivity, and the large number of brain connections leading to spurious findings.  In this paper, we propose a novel blind source separation method with low-rank structure and uniform sparsity (LOCUS) as a fully data-driven decomposition method for network measures. Compared with the existing method that vectorizes connectivity matrices ignoring brain network topology, LOCUS achieves more efficient and accurate source separation for connectivity matrices using low-rank structure. We propose a novel angle-based uniform sparsity regularization that demonstrates better performance than the existing sparsity controls for low-rank tensor methods. We propose a highly efficient iterative node-rotation algorithm that exploits the block multi-convexity of the objective function to solve the non-convex optimization problem for learning LOCUS. We illustrate the advantage of LOCUS through extensive simulation studies. Application of LOCUS to Philadelphia Neurodevelopmental Cohort neuroimaging study reveals biologically insightful connectivity traits which are not found using the existing method.

\end{abstract}


\end{frontmatter}

\section{Introduction}

 In recent years, network-oriented analyses have become an important research field in neuroscience for understanding brain organization and its involvement in neurodevelopment and mental disorders \citep{bullmore2009complex, deco2011emerging, satterthwaite2014linked, kemmer2015network,wang2016efficient,wang2019hierarchical}. In neuroimaging studies, network measures are derived from various neuroimaging modalities to reflect functional or structural connections between a set of nodes or brain regions. The network measures are typically encoded as symmetric matrices where the entries represent brain connectivity between pairs of nodes or regions in the brain. For example, derived from functional magnetic resonance imaging (fMRI) or electroencephalogram (EEG), functional connectivity (FC) measures the dependence between temporal dynamics in neural processing of spatially disjoint brain regions \citep{biswal1995functional,lang2012brain,kemmer2018evaluating,kundu2019novel}. Some commonly used FC measures include the Pearson correlation,  partial correlation, mutual information and coherence \citep{smith2011network,church2008control,seeley2009neurodegenerative,wang2016efficient}. Brain connectivity matrices consisted of connectivity measures contain important information to understand brain organization and its changes due to neurodevelopment, disease progression or treatment  \citep{finn2015functional,amico2018quest}. 

\begin{figure}
    \centering
    \includegraphics[scale =.45]{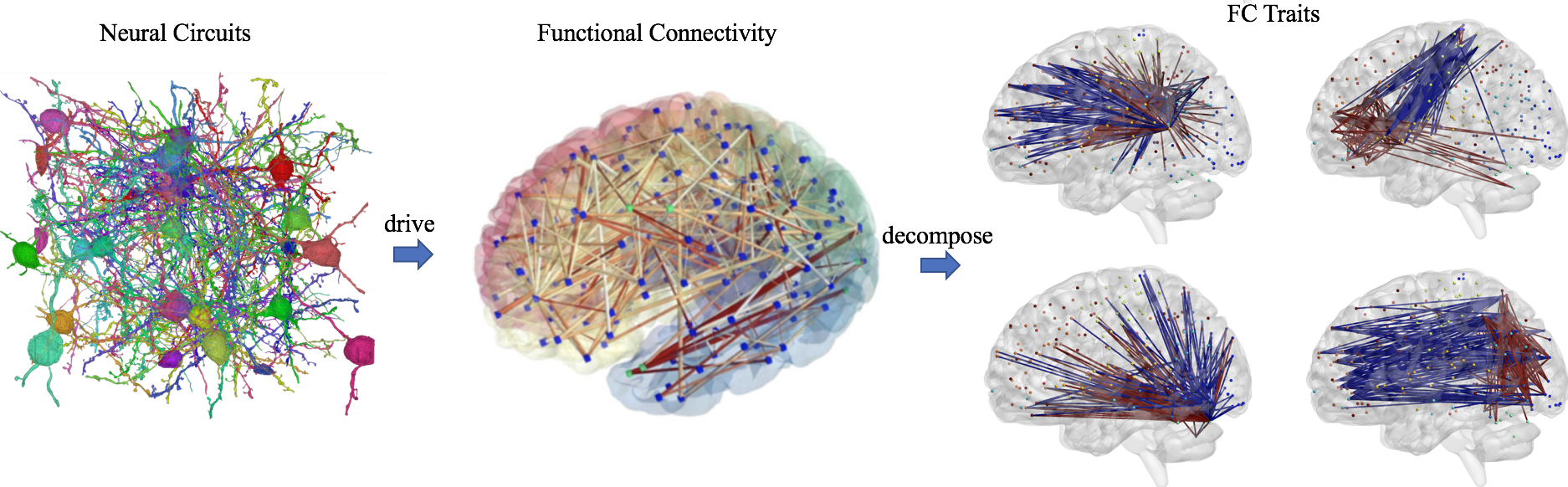}
    \caption{Illustration of decomposing the brain functional network into various underlying neurocircuitry traits.}
    \label{fig:idea_decomFC}
\end{figure}

There are several major challenges in network analysis in neuroimaging studies.  First, to fully capture the whole brain organization, connectivity matrices are usually high dimensional \citep{chung2018statistical}. Voxel-level brain network based on fMRI data contains tens of thousands of nodes and nearly half a billion edges. To reduce the network dimension,  atlas-based brain networks are commonly constructed based on a brain atlas or node system such as the automated anatomical labeling (AAL) atlas \citep{tzourio2002automated} and the more recent Power's node system \citep{power2011functional} and Glassier's atlas \citep{glasser2013minimal}. Although the number of nodes are reduced dramatically in these atlas-based networks, there are still hundreds of nodes and hundreds of thousands of edges  \citep{chung2018statistical,solo2018connectivity,wu2013mapping,wang2016efficient}. The high dimensionality makes it challenging to  reliably estimate brain networks for scientific discoveries. Secondly, the brain connectome is a complex organization encompassing many underlying neural circuits. The observed connectivity matrix, measuring the overall connectivity patterns across the brain, represents aggregated information from various underlying neural circuits (Figure \ref{fig:idea_decomFC}). Currently, there is a lack of methods to reliably decompose observed connectivity matrices to recover the underlying neurocircuitry. Neuroscience literature has found that different neural circuits develop at different age and rates during early brain development \citep{hoff2013development}. Studies have also found that demographic- or disease-related alterations in brain network usually occur in certain neural circuits instead of in the whole brain connectome \citep{mayberg2003modulating,williams2016precision}.  Without consistent dissection of neural circuits, it is often not clear which neural circuits are primary drivers of neurodevelopmental changes, subpopulation differences and clinical symptoms.  Thirdly, given the large number of edges present in the brain networks, there is a high possibility of spurious findings in terms of neurodevelopmental- and disease-related differences in brain connections across the network in neuroscience research. 

To address the aforementioned challenges, we aim to develop a statistical blind signal separation framework  for decomposing imaging-based brain connectivity data to reveal underlying neural circuits. The research is motivated by Philadelphia Neurodevelopmental Cohort (PNC) study. The PNC study is a large-scale neurodevelopmental study aiming to understand brain maturation and development of cognition \citep{satterthwaite2014linked} during late childhood and adolescent. The PNC study includes a population-based sample of participants aged 8–21 years in the greater Philadelphia area. 
Brain imaging were acquired for a subset of the participants. In particular, resting-state fMRI (rs-fMRI) was acquired to study intrinsic brain organization related to spontaneous neural activity. One research goal is to investigate how brain functional connections are altered during neurodevelopment and whether there are major differences in brain organization between sub-populations such as gender groups. Given the complexity of brain organizations and previous finding suggesting different brain networks develop at various age and paces during brain maturation, it is important to obtain reliable dissection of neural circuits in order to investigate varying neurodevelopment in circuits. 

Currently, one of the most popular blind source separation methods for decomposing neuroimaging data is independent component analysis (ICA). In neuroscience research, ICA is widely used  for dimension reduction, denoising and extraction of latent neural components. ICA has achieved great success in a range of neuroimaging applications \citep{beckmann2004probabilistic,guo2011general,shi2016investigating,contreras2017cognitive,mejia2019template,wang2019hierarchical,lukemire2020hint}. However, existing ICA applications have mainly focused on decomposing observed neural activity signals such as the blood oxygen level-dependent (BOLD) series from fMRI or the electrodes signal series from EEG. Brain connectivity data have different properties as compared to activity data. For example, a connectivity matrix is typically symmetric and its diagonal elements (i.e. self-connections) are often not of interest. Therefore, the relevant information in a connectivity matrix is captured by its lower or upper triangle matrix \citep{amico2017mapping}. Traditional matrix decomposition approaches such as ICA cannot be directly applied to brain connectivity matrices. Recently,  \citet{amico2017mapping} proposed a connectivity independent component analysis framework (connICA). connICA first vectorizes the connectivity matrices and then utilizes existing ICA algorithms for decomposition.  Ignoring the dependence structure across edges in the brain network, connICA method treats each edge independently and has a large number of parameters, which reduces accuracy and reliability in extracting connectivity components. connICA does not include sparsity regularization either, leading to noisy estimates in connectivity analysis. When applying connICA to decompose the resting-state functional connectivity data from PNC study, one obtains highly dense connectivity traits which increase the risk of spurious findings. Due to the noisy estimates, connICA generates less reproducible neural circuits and fails to discover several underlying connectivity traits that are associated with individuals' age and gender (see Section \ref{realsec}). Results of connICA in the PNC study reveal the limitation of the existing blind source separation method for brain connectivity.

In recent years, there have been methods development in using tensor or low-rank factorization for studying network data in neuroimaging analysis \citep{durante2017nonparametric, zhou2013tensor,sun2017store,wang2017bayesian,eavani2015identifying}. The existing methods mainly focus on using the low-rank structure to reduce the size of the high-dimensional parameters in tensor regressions or to model binary brain networks. Specifically, existing tensor regression methods in brain network analysis \citep{zhou2013tensor,sun2017store} either treat brain connectivity matrices as a high dimensional predictor or model it as a tensor outcome. The low-rank structure is assumed for the regression parameters to efficiently learn the high-dimensional parameters in regression setting. In some other work, Bayesian methods are developed for binary brain networks obtained by thresholding the observed brain connectivity measures \citep{durante2017nonparametric, wang2017bayesian}. These papers mainly adopt low-rank factorization in modeling the parameters characterizing the probability mass function of the binary network data. The existing methods are not designed to conduct fully data-driven decomposition of connectivity data to extract latent connectivity traits corresponding to various neural circuits. \citet{eavani2015identifying} proposes to approximate brain correlation matrices with a non-negative sum of a set of rank one matrices. Though this method is applicable to connectivity data, the constraint that each of the component has to be a rank one matrix limits its ability to recover neural circuits that often have topological structures and spatial patterns beyond a rank one structure. Other relevant work includes subspace learning dimension reduction methods for brain network analysis \citep{wang2019common} which mainly target using reduced subspace to estimate covariance or precision matrices. 

\begin{figure}
\begin{center}
\includegraphics[scale=0.38]{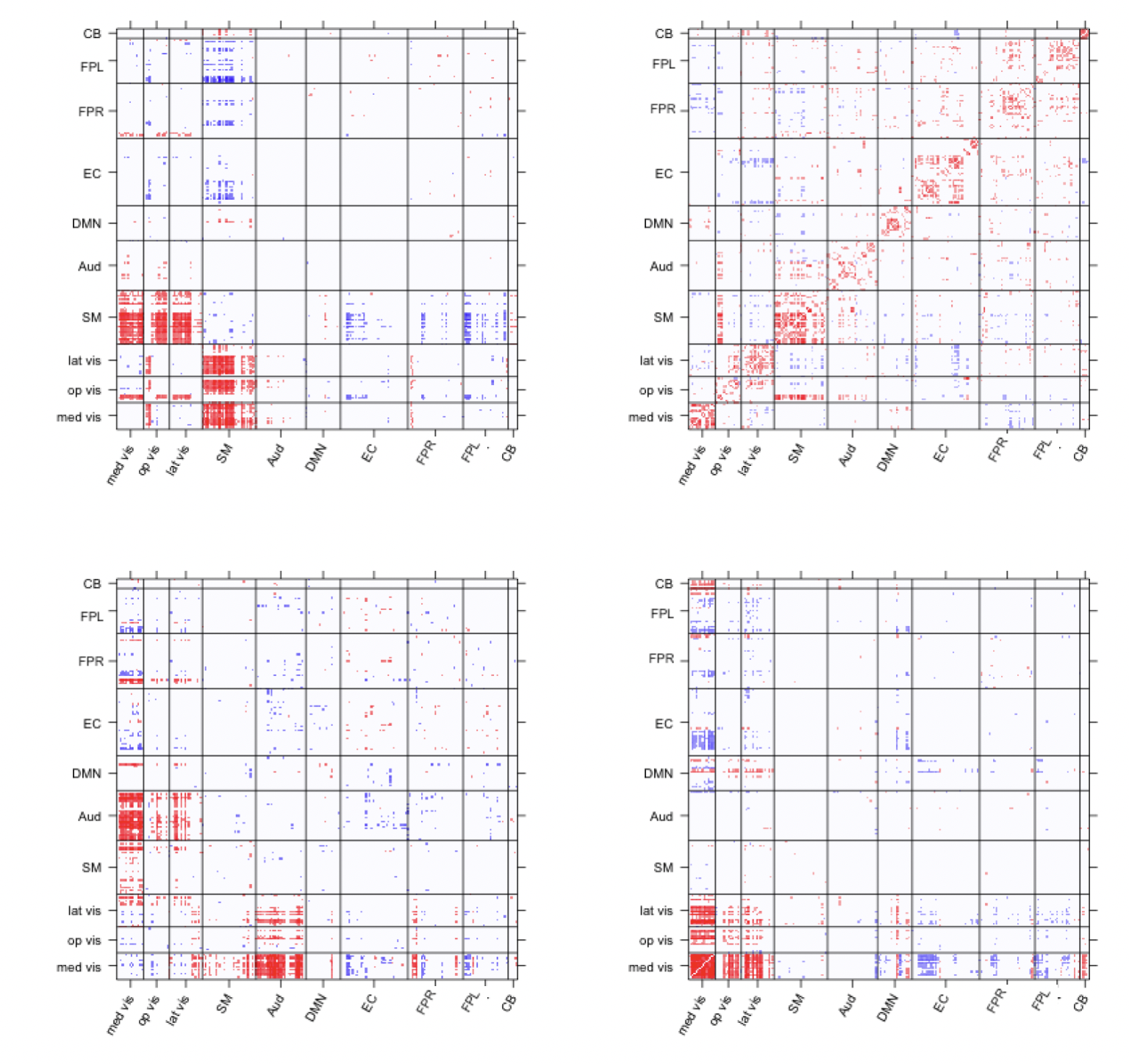}
\caption{\small{Example connectivity traits extracted from resting-state fMRI data.}}\label{fig1:illustrate2}
\end{center}
\end{figure}

In this paper, we present a novel {\it lo}w-rank decomposition of brain {\it c}onnectivity with {\it u}niform {\it s}parsity (LOCUS) method for decomposing imaging-based brain network measures to identify underlying source signals characterizing connectivity traits. LOCUS is a fully data-driven blind source separation method for decomposing brain connectivity data derived from various network measures.  Specifically, LOCUS decomposes subjects' connectivity data, $\bm Y$, into a linear combination of latent connectivity traits or source signals, $\{\bm S_\ell\}_{\ell=1}^{q}$, weighted by mixing coefficients $\{\bm{a}_\ell\}_{\ell=1}^{q}$, i.e. $\bm Y = \sum_{\ell=1}^{q} {\bm{a}_\ell} \bm S_\ell +$ error. Here, each of the connectivity source signals $\bm S_\ell$ represents an underlying neural circuit and the mixing coefficients $ \bm{a}_\ell$ represent subject-specific loadings on the trait. We propose to model the source signals $\bm S_\ell$ using a low-rank structure $\bm X_{\ell}\bm D_{\ell}\bm X_{\ell}'$ where $\bm D_{\ell}$ is a diagonal matrix. This is well motivated by the observation that brain connectivity traits often have block-diagonal or banded structure (Figure \ref{fig1:illustrate2}) that can be efficiently captured with a low-rank factorization \citep{zhou2013tensor}. The low-rank structure leads to a significant reduction in the number of parameters, hence improving accuracy and reliability in the recovery of underlying connectivity traits. Compared to \citet{eavani2015identifying} which restricts each component matrix to be rank one, we propose an adaptive rank selection approach to flexibly choose the rank for each of the connectivity source signals. The source-specific rank selection allows better accommodation of varying spatial patterns in the neural circuits across the brain.  Moreover, the subject-specific trait loadings generated by LOCUS quantify the prominence or presence of each trait in a subject’s connectivity. These subject-specific trait loadings capture between-subject heterogeneity in each of the connectivity traits and also allow identifying which traits are associated with clinical or demographical variables. The subject-specific trait loadings and connectivity trait-specific association analysis are not available from some other network modeling methods \citep{wang2019common}.

%To better capture varying spatial patterns in the neural circuits across the brain, we propose an adaptive rank selection approach to choose the rank for each of the connectivity source signals.

Moreover, to reduce spurious scientific findings, it is necessary to consider appropriate regularization method to achieve the sparsity in the extracted latent source signals, which have low-rank structure.  Currently, there are two major approaches of sparsity penalization for low-rank factorization \citep{raskutti2015convex,wang2018sparse}: the vector-wise sparsity constraint that aims to achieve element-wise sparsity of the column vectors of $\bm X_{\ell}$ \citep{zhou2013tensor,eavani2015identifying,sun2017store,li2018tucker}, and the low-rank matrix constraint that aims to recover a low-rank matrix via regularization on $\bm D_{\ell}$ such as minimizing the nuclear norm of $\bm S_{\ell}$ \citep{chen2013reduced,rabusseau2016low,fan2017generalized, yuan2016tensor}. Both of the existing sparsity controls have limitations for achieving our goal which is to recover parsimonious connectivity traits $\bm S_{\ell}$. The low rank constraint that aims to achieve low rankness in $\bm S_{\ell}$ does not necessarily ensure the sparsity in $\bm S_{\ell}$ \citep{sun2017store}. The vector-wise sparsity control on $\bm X_{\ell}$ often generates structured noises
in estimating the connectivity traits given that $\bm S_{\ell}$ depends on the inner products of the vectors.  

Since our main interest is to derive sparse connectivity traits, we propose a more intuitive way of sparsity regularization by directly penalizing on latent sources $\bm S_{\ell}$ via the low-rank structure $\bm X_{\ell}\bm D_{\ell}\bm X_{\ell}'$. Later in the paper, we will show it is essentially as angle-based penalization. Through extensive simulation studies, we show this new sparsity method demonstrates more accurate and robust performance in recovering underlying traits than the commonly used sparsity controls. Furthermore, compared with some existing sparsity controls which require numerical methods to solve, our new sparsity penalization allows for explicit analytic solutions to the optimization function in the estimation, which increases computational efficiency. The learning of LOCUS is formulated as a non-convex optimization problem. We show that the optimization function has a block multi-convex structure \citep{gorski2007biconvex}. We develop an efficient node-rotation algorithm with closed-form solutions at each iteration for estimating the parameters in LOCUS.

In summary, we proposed a novel blind source separation method for decomposing brain connectivity matrix.  Our contributions include following. First, LOCUS provides a formal signal separation approach with a low-rank structure for decomposing brain connectivity matrices. The low-rank structure and our adaptive rank selection method leads to considerable reduction in parameters and improved flexibility and accuracy in recovering latent connectivity traits. These advantages allow LOCUS to achieve more efficient and reliable source separation for connectivity metrics. Secondly, we propose a novel sparsity regularization method that aims to control the element-wise or uniform sparsity on the connectivity source matrix reconstructed from the low-rank factorization. Compared with the commonly used sparsity  methods that focus on controlling sparsity on the vector or diagonal matrix components in the low-rank factorization, our method aims to control the sparsity in the overall matrix reconstructed from the low-rank structure, which directly targets the output, i.e. connectivity trait, that we are interested in. This new sparsity control has shown highly promising results in the simulation studies and provide a new type of sparsity regularization that can be generally applied in low-rank structure involved models such as tensor regressions or covariance modeling. Thirdly, we establish the block multi-convexity in LOCUS objective function and develop an efficient node-rotation algorithm. The proposed model and the estimation algorithm demonstrate superior performance in recovering the underlying source signals though our extensive simulation studies.  

The overall structure of the paper is organized as follows. Section \ref{sec2} introduces the methodology including model specification and the estimation algorithm. Section \ref{realsec} applies LOCUS to the PNC rs-fMRI connectivity data to investigate neurodevelopment in neural circuits. Section \ref{sec3} demonstrate the performance of LOCUS in comparison with other methods via simulation studies. Section \ref{Discussion} is for discussion and conclusion.

\section{Methodology}\label{sec2}

\subsection{Notations and Structure}\label{notation}

We define vector norm $\|\bm x\|_h = (\sum_i |x_i|^h)^{\frac{1}{h}}$ with an integer $h\geq 1$, and we denote the Frobenius norm of a matrix by $\|\bm X\|_F = (\sum_{ij}X_{ij}^2)^{1/2}$. Suppose we observe brain connectivity data from $N$ subjects. Let $\bm Y_{i}$ denote a $V \times V$ symmetric brain connectivity matrix for the $i$th ($i=1,\ldots, N$) subject with $V$ denoting the number of nodes, ${\bm Y}_{i}(u,v) \in \mathcal{R}$ representing the strength of connection between node $u$ and $v$ in the brain. %$ (u,v \in \{1, \ldots,V\})$. 
${\bm Y}_{i}$ is obtained by performing proper transformations on brain connectivity measures. Since the diagonal of ${\bm Y}_{i}$ which represents self-relationship in the network is typically not of interest, we define a vector $\bm y_i$ based on the upper triangular elements of ${\bm Y}_{i}$, i.e.  $\bm y_i=\mathcal{L}({\bm Y}_{i})$ where $\L({\bm Y}_{i}) = [{\bm Y}_{i}(1,2),{\bm Y}_{i}(1,3),...,{\bm Y}_{i}(V-1,V)]'$. Here, $\L: \mathcal{R}^{V\times V} \rightarrow \mathcal{R}^{p}$ with $p= V(V-1)/2$.

\subsection{Model}\label{LOCUS}
In this section, we introduce the LOCUS framework for decomposing multi-subject connectivity matrices. 

\subsubsection{The LOCUS Decomposition Model}
 We propose the following LOCUS model to decompose the multi-subject connectivity matrices to extract latent connectivity sources. Motivated by the observed patterns of brain connectivity traits which often have block-diagonal or banded structure (Figure \ref{fig1:illustrate2}), we model connectivity sources with a low-rank structure which can considerably reduce the number of parameters while capturing the network characteristics. Specifically, 

\vspace{-0.02in}
\begin{equation}
\label{eq:LocusICA1}
\bm y_i=\sum_{\ell=1}^{q}a_{i\ell}\bm s_{\ell}+\bm e_i, 
\end{equation}
\vspace{-0.02in}
\noindent where $\bm s_\ell = \L(\bm S_{\ell}) \in \mathcal{R}^{p}$ $(\ell=1,\ldots,q)$ is the source signal of the $\ell$th connectivity source or trait and we assume independence across the $q$ traits. A connectivity trait represents a set of  between-region brain connections that tend to occur together. $\{a_{i\ell}\}$ are the mixing coefficients or trait loadings which mixes the traits to generate the observed connectivity, $\bm e_i \in \mathcal{R}^{p} $ is an error term independent of the source signals. The number of latent sources, i.e. $q$, can be determined using methods such as the Laplace approximation \citep{minka2000automatic} or based on the reproducibility and interpretability of the extracted latent sources. We can also rewrite the LOCUS model in (\ref{eq:LocusICA1}) across subjects as,
\begin{equation}
\label{eq:LocusICAgroup}
    \bm Y=\bm A \bm S+\bm E,
\end{equation}
where $\bm Y = [\bm y_1, ... ,\bm y_N]' \in \mathcal{R}^{N \times p}$
is the multi-subject connectivity data, 
$\bm S =[\bm s_1,...,\bm s_q]'\in \mathcal{R}^{q \times p}$ is the connectivity traits matrix, 
$\bm A=\{a_{i\ell}\} \in \mathcal{R}^{N \times q}$ is the mixing/loading matrix, and $\bm E= [\bm e_1, ... ,\bm e_N]' \in \mathcal{R}^{N \times p}$. 

\begin{figure}
    \centering
    \includegraphics[scale = 0.65]{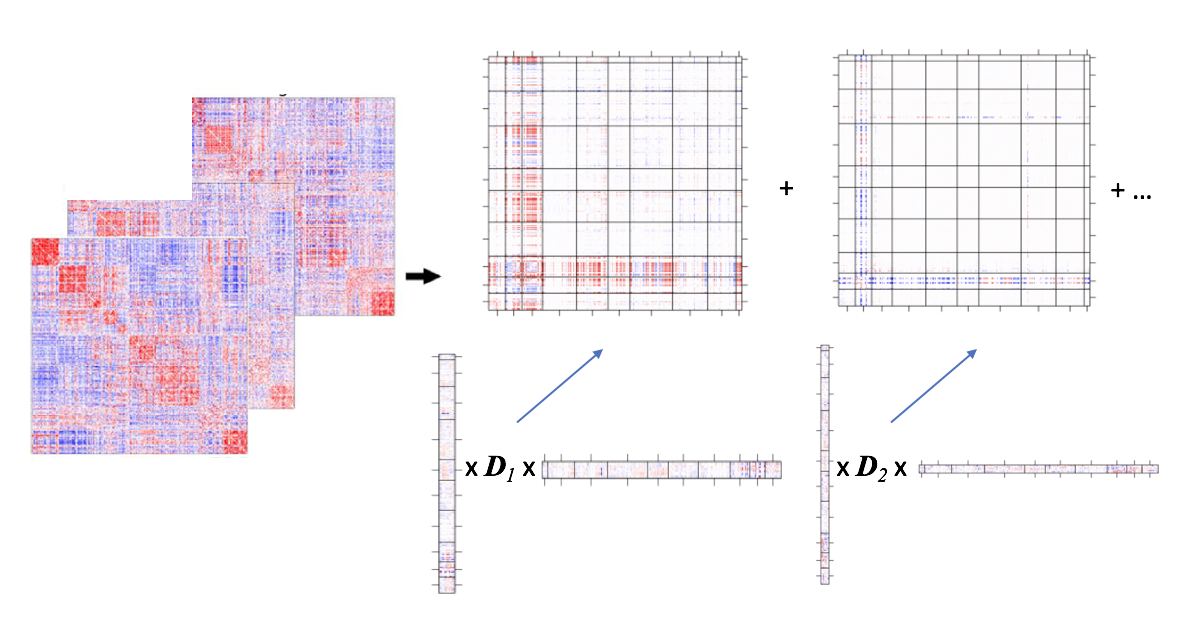}
    \caption{Illustration of LOCUS framework for decomposing subjects' brain functional connectivity (FC) matrices into latent source signal matrices representing various underlying neurocircuitry traits. Each latent source signal matrix is further modeled via a symmetric low-rank factorization. }
    \label{fig:idea_locus}
\end{figure}

We model the connectivity source signals via a low-rank structure, i.e. 
\begin{equation}
\label{eq:loc}
\bm s_{\ell} = \L(\bm X_\ell\bm D_\ell\bm X_\ell'),
\end{equation}
where $\bm X_\ell=[\bm x_\ell^{(1)},\ldots,\bm x_{\ell}^{(R_\ell)}] \in \mathcal{R}^{V\times R_{\ell}}$ with $R_{\ell}< V$ and each column $\bm x_{\ell}^{(r)}$ ($r=1,\ldots,R_\ell$) is a $V\times 1$ vector with unit norm, i.e. $\|\bm x_{\ell}^{(r)}\|_2 = 1$ for identifiability purpose. $\bm D_\ell$ is a diagonal matrix with diagonal elements $\bm d_{\ell}= (d^{(1)}_{\ell},..,d^{(R_\ell)}_{\ell})$. The low-rank structure implies the $V$ nodes reside in a reduced subspace of $R_\ell$ dimensions, i.e. $\bm s_\ell=\L (\sum_{r=1}^{R_{\ell}} d_{\ell}^{(r)} \bm x_{\ell}^{(r)}\bm x_{\ell}^{(r)'})$ where the $r$th column $\bm x_{\ell}^{(r)} \in \mathcal{R}^{V\times 1}$ represents the coordinates of the $V$ nodes in the $r$th dimension and $d^{(r)}_{\ell} $ reflects the contribution of the $r$th dimension in generating $\bm s_\ell$. Each row of $\bm X_\ell$, i.e.  $\bm x_\ell(v)'$ with $\bm x_\ell(v) \in \mathcal{R}^{R_{\ell}\times 1}$, represents the coordinates of the $v$th node in the $R_\ell$ dimensional latent subspace. As shown in Figure \ref{fig1:illustrate2}, the network property and topological structures may vary considerably across different connectivity traits. Therefore, we let the subspace rank $R_{\ell}$ to be specific to each latent source to accommodate such differences. Figure \ref{fig:idea_locus} illustrates the framework of LOCUS method.

The decomposition part of the LOCUS model in (\ref{eq:LocusICA1}) takes a similar form as the probabilistic ICA \citep{hyvarinen2001independent, beckmann2004probabilistic} which is a classical blind source separation method. However, there are key distinctions between LOCUS and ICA methods. For example, the existing connectivity ICA methods such as connICA \citep{amico2017mapping} simply vectorizes a connectivity matrix and assumes the $p$ elements in $\bm s_{\ell}$ are independent random samples from the same latent variable, ignoring the network topology structure across the brain. In comparison, LOCUS models the source signal using a low-rank structure which is well-motivated by the characteristics of the connectivity metrics. The low-rank structure offers several major advantages. First, it leads to a significant reduction of the number of parameters from a quadratic function of the number of nodes $V$ to a linear function of $V$, hence improving reliability in estimation. Furthermore, the low-rank structure offers appealing interpretations for neurocircuitry structures. Each of the $R_\ell$ subspace dimension can potentially reveal an underlying neurophysiological event that contributes to the connectivity trait where $\bm x_{\ell}^{(r)}$ captures the activity of $V$ nodes in the $r$th neural event, allowing us to identify key nodes involved in a connectivity trait. The connection between two nodes in the $\ell$th trait depends on the similarity in the activity of the nodes across the ${R_{\ell}}$ neural events, i.e. $\bm S_{\ell}(u,v) = \bm x_{\ell}(u)'\bm D_{\ell} \bm x_{\ell}(v)$, which coincides with the definition of brain connectivity in neuroscience \citep{friston1993functional, friston2011functional}. LOCUS reveals underlying neural events and brain nodes contributing to the connectivity traits. The existing ICA methods do not provide such interpretations and insights.

\subsubsection{A novel sparsity regularization}\label{uni-spar}

In LOCUS, we model each latent source $\bm S_\ell$ with a low-rank factorization.  To obtain parsimonious results for the connectivity traits, we propose a new sparsity method that aims to achieve element-wise sparsity on the reconstructed connectivity traits based on the low-rank structure. Currently, the commonly adopted sparsity regularization methods in low-rank factorization and tensor decomposition methods fall into two major categories: the vector-wise sparsity constraint or low-rank matrix constraint \citep{raskutti2015convex,wang2018sparse}. The vector-wise sparsity methods aim to achieve element-wise sparsity of the column vectors $\{\bm x_\ell^{(r)}\}$ by minimizing a penalization terms based on the L1 or L2 norms of $\{\bm x_\ell^{(r)}\}$ \citep{zhou2013tensor, li2018tucker, fan2001variable,zhang2010nearly, eavani2015identifying}. The low-rank constraint methods aim to recover a low-rank matrix on $\bm S_{\ell}$ such as minimizing the nuclear norm of $\bm S_{\ell}$ \citep{chen2013reduced,rabusseau2016low,fan2017generalized,yuan2016tensor}. In LOCUS, our goal is to recover parsimonious connectivity traits $\bm S_{\ell}$. The existing vector-wise sparsity control and low rank constraint methods have certain limitations in achieving this goal. The vector-wise sparsity control, which aims to achieve sparsity in the column vectors $\{\bm x_\ell^{(r)}\}$, may not lead to sparsity in the connectivity traits which depends on the dot product between the vectors. The nuclear norm method that aims to achieve low rankness in $\bm S_{\ell}$ does not necessarily lead to element-wise sparsity in $\bm S_{\ell}$.

We propose a novel sparsity regularization that aims to achieve sparsity on the reconstructed connectivity trait matrix based on the low rank structure. i.e. $\bm s_\ell= \L(\bm X_\ell \bm D_\ell\bm X_\ell')$. The objective function for estimating the LOCUS model with the new sparsity control is defined as follows,
\begin{equation}
\label{eq:locus_lp}
\text{min} \sum_{i=1}^N\| \bm y_i - \sum_{\ell=1}^q a_{i\ell} \L(\bm X_\ell\bm D_\ell\bm X_\ell') \|_2^2 + \phi\sum_{\ell=1}^q \sum_{u<v} |\bm x_{\ell}(u)'\bm D_\ell\bm x_{\ell}(v)|,
\end{equation}
where $\phi$ is a tuning parameter for the sparsity term. The penalty term in (\ref{eq:locus_lp}) aims to achieve element-wise or uniform sparsity on the upper triangular elements of the reconstructed connectivity traits, i.e. $\bm S_{\ell}(u,v)=\bm x_{\ell}(u)'\bm D_\ell\bm x_{\ell}(v)$ across all node pairs $(u,v)$ with $u<v$. We denote this novel regularization method as L1-based uniform sparsity across connections. Alternative penalization functions such as L2 regularization, SCAD or MCP can also be adopted for our sparsity regularization. 

% we listed following two models (\ref{eq:locus_l1}), (\ref{eq:locus_nulear}) with existing penalization methods to better illustrate the distinction between the proposed penalization method.  

%\begin{equation}
%\label{eq:locus_l1}
%\text{min} \sum_{i=1}^N\| \bm y_i - \sum_{\ell=1}^q a_{i\ell} \L(\bm X_\ell\bm D_\ell\bm X_\ell') \|_2^2 + \phi\sum_{\ell=1}^q \sum_{v = 1}^V \|\bm x_{\ell}(v)'\|_1,
%\end{equation}

%\begin{equation}
%\label{eq:locus_nulear}
%\text{min} \sum_{i=1}^N\| \bm y_i - \sum_{\ell=1}^q a_{i\ell} \L(\bm S_\ell) \|_2^2 + \phi\sum_{\ell=1}^q \|\bm S_\ell\|_*,
%\end{equation}

%where $\|*\|_*$ denotes nuclear norm. 

To provide insights on the proposed sparsity method, we note that the penalization term in (\ref{eq:locus_lp}) is essentially an angle-based penalization. Specifically, for the $\ell$th latent source, the penalty term aims to minimize the summation of the weighted inner products, i.e. angles,  between latent coordinates vectors between all the node pairs in the brain. Since our endpoint of interest in LOCUS is the connectivity between the nodes, the proposed sparsity regularization based on the angles, which captures the dependence between the node pairs, is intuitively and theoretically appealing. Additionally, we show in the following that with this proposed sparsity term we can develop an efficient learning algorithm with closed-form solutions in each updating step to optimize the objective function. This is another advantage of the our sparsity regularization compared with some existing sparsity methods that require numerical methods (i.e. gradient method) to solve.

\subsection{Estimation}\label{LOCUS-algorithm}
In this section, we present the estimation method for learning the parameters in LOCUS. First, we introduce the data preprocessing step prior to LOCUS. We then present an efficient node-rotation algorithm and show block multi-convexity for the proposed optimization function. We also propose the procedure for tuning parameter selection including an adaptive selection approach for choosing the source-specific rank parameters.  

\subsubsection{Preprocessing prior to LOCUS decomposition}
Prior to LOCUS decomposition, we take several preprocessing steps that are commonly adopted in blind source separation, which includes centering, dimension reduction and whitening. The preprocessing is generally performed to facilitate the subsequent decomposition by reducing the computational load and avoid overfitting \citep{hyvarinen2001independent}. Following the preprocessing procedure from previous work \citep{beckmann2004probabilistic,  shi2016investigating, wang2019hierarchical}, we first demean the group connectivity data $\bf Y$ and then perform a dimension reduction and whitening procedure on the demeaned data. That is, $\widetilde{\bY} = \bm H \bY $, where $\bm H =  (\bm \Lambda_q - \widetilde{\sigma}_q^2 \bm I)^{-1/2} \bm U_q'$. $\bm U_q$ and $\bm \Lambda_q$ contains the first $q$ eigenvectors and eigenvalues based on singular value decomposition of $\bm Y$. The residual variance, $\widetilde{\sigma}_q^2$, represents the variability in $\bf Y$ that is not explained by the extracted $q$ latent sources and is estimated by the average of the smallest $N - q$ eigenvalues in $\bm \Lambda_q$. The preprocessed data $\widetilde{\bY} $ is of dimension $q \times p$ where each column corresponds to one of the $p$ connections in the brain. 

With the preprocessing, the model in (\ref{eq:LocusICAgroup}) can be re-expressed on the reduced and sphered space as following:
\begin{equation}
\label{eq:LocusICAgroup_preproc}
   \widetilde{\bY}=\widetilde{\bm A} \bm S + \widetilde{\bm E},
\end{equation}
where $\widetilde{\bm A}=\bm H \bm A$ and $\widetilde{\bm E}=\bm H \bm E$. Due to the whitening in the preprocessing, the mixing matrix on the reduced space $\widetilde{\bm A} = \{\widetilde{a}_{i\ell}\} \in \mathcal{R}^{q \times q}$  is orthogonal \citep{hyvarinen2000independent, beckmann2004probabilistic}.
Note that the dimension reduction in the preprocessing is performed on the row space of $\bf Y$ which corresponds to the subject domain and does not affect the column space of $\bf Y$ which corresponds to the connectivity domain. The optimization function for LOCUS with the preprocessed data is then
\begin{equation}\label{eq:locus_preprocess}
 \min\limits_{\widetilde{\bm A}, \{\bm X_\ell, \bm D_{\ell}\}} \sum_{i=1}^q\| \widetilde{\bm y}_i - \sum_{\ell=1}^q \widetilde{a}_{i\ell} \mathcal{L}(\bm X_\ell \bm D_\ell \bm X_\ell')  \|_2^2 + \phi\sum_{\ell=1}^{q}\sum_{u<v} |\bm x_\ell(u)'\bm D_\ell\bm x_\ell(v)|,
\end{equation}
where $\widetilde{\bm y}_i$ is the transpose of the $i$th row in $\widetilde{\bY}$ for $i=1,\ldots,q$. 

The following Lemma presents a way to further rewrite the optimization function to facilitate the estimation.

\begin{lemma0}\label{equ_prcess}
With an orthogonal mixing matrix $\widetilde{\bm A}$, the optimization in (\ref{eq:locus_preprocess}) is equivalent to 
\begin{equation}\label{eq:locus_preprocess2}
    \min\limits_{\widetilde{\bm A}, \{\bm X_\ell, \bm D_{\ell}\}} \sum_{\ell=1}^q\| \widetilde{\bY}'\widetilde{\bm a}_{\ell} -  \mathcal{L}(\bm X_{\ell} \bm D_{\ell} \bm X_{\ell}')  \|_2^2 + \phi\sum_{\ell=1}^{q}\sum_{u<v} |\bm x_{\ell}(u)'\bm D_{\ell}\bm x_{\ell}(v)|,
\end{equation}
where $\widetilde{\bm a}_{\ell}$ is the ${\ell}$th column of $\widetilde{\bm A}$. 
\end{lemma0}
The proof of the Lemma is presented in Appendix. The rewriting in Lemma \ref{equ_prcess} is performed on the first term in the optimization function which evaluates the difference between the preprocessed connectivity data and the LOCUS model fit. The rewriting essentially changes the first term from evaluating the difference on the processed data domain in (\ref{eq:locus_preprocess}) to evaluating the difference on the source signal domain in (\ref{eq:locus_preprocess2}). Since the preprocessed data are whitened and reduced to the same dimension as the source signals, we can show that the difference on these two domains are equivalent (see Appendix for proof of Lemma \ref{equ_prcess}). The change to the source signal domain allows us to develop an efficient updating algorithm on each of the latent sources in the estimation as shown in the following section.

\subsubsection{A Node-Rotation Algorithm}
 We propose an efficient algorithm to solve the optimization problem in (\ref{eq:locus_preprocess2}). Denote $\bm \Theta=[\widetilde{\bm A}, \{\bm X_\ell, \bm D_{\ell}\}]$ as the parameters to learn. We propose the following iterative estimation algorithm which includes three major updating steps. A summary of the algorithm is presented in Algorithm~\ref{locus_alg}. 

{\bf Step 1}: {\it Updating $\bm X_{\ell}$}. We propose a novel node-rotation algorithm that updates $\bm X_\ell$ at one of the node $v$ while conditioning on the rest of the nodes and then rotating across the nodes. This algorithm exploits the conditional convexity in $\bm x_\ell(v)$ given the other parameters. Specifically, at the $t$th iteration, we update $\hat{\bm x}_\ell^{(t)}(v)$, $v = 1,..,V$, conditioning on $\widetilde{\bm A}$, $\bm D_{\ell}$ and $\bm X_{\ell}(-v)$ estimated from the $t-1$ iteration, where $\bm X_{\ell}(-v)$ is $\bm X_{\ell}$ with the $v$th row removed. The updated $\hat{\bm x}_\ell^{(t)}(v)$ is obtained via the following,
\begin{align} \label{update2}
     \min\limits_{\bm x_{\ell}(v)} \textbf{ } \Big\| \widetilde{\bm Y}_{\{v\}}'\hat{\widetilde{\bm a}_{\ell}} - \hat{{\bm X}}_{\ell}(-v) {\hat{\bm D}}_{\ell} \bm x_{\ell}(v)\Big\|_2^2 + \phi \sum_{\substack{u=1 \\ u\ne v}}^V | {\hat{\bm x}_{\ell}}(u)' {\hat{\bm D}_{\ell}} \bm x_{\ell}(v)  |,
\end{align}
where $\widetilde{\bm Y}_{\{v\}}$ is a $q \times (V-1)$ sub-matrix of $\widetilde{\bm Y}$ which includes the subset of columns in $\widetilde{\bm Y}$ that correspond to connections involving node $v$. A detailed derivation from (\ref{eq:locus_preprocess2}) to (\ref{update2}) is provided in the Appendix. It is straightforward to show that the optimization in (\ref{update2}) is convex.

We propose the following procedure for solving (\ref{update2}). First, we define $\bm b_{\ell\{v\}} \in \mathcal{R}^{V-1}$ to represent $\hat{{\bm X}}_{\ell}(-v) {\hat{\bm D}}_{\ell} \bm x_{\ell}(v)$. We obtain estimate $\hat{\bm b}_{\ell\{v\}}$ via the optimization below which is a rewrite of (\ref{update2}),
\begin{equation}
\label{eq:two-step1}
    \min\limits_{\bm b_{\ell\{v\}} \in \mathcal{R}^{V-1}} \Big\|\widetilde{\bm Y}_{\{v\}}'\widetilde{\bm a}_{\ell}   -  \bm b_{\ell\{v\}} \Big\|_2^2 + \phi \Big\|\bm b_{\ell\{v\}} \Big\|_1,
\end{equation}
Following \citet{fan2001variable}, we derive the following analytic solution for (\ref{eq:two-step1}),
\begin{align} 
\hat{\bm b}_{\ell\{v\}}= \text{diag}\Big(\text{sgn}\big(\widetilde{\bm Y}_{\{v\}}'\widetilde{\bm a}_\ell\big)\Big) \delta\big( |\widetilde{\bm Y}_{\{v\}}'\widetilde{\bm a}_\ell| - \frac{\phi}{2} \bm 1_{V-1} \big), \nonumber 
\end{align}
where sgn represents sign function for each element and $\delta$ denotes a rectifier function ($\delta(x) = x$ if $x > 0$ otherwise 0). 
 
In the next step, we project $\hat{\bm b}_{\ell\{v\}}$ to the low-rank space spanned by $\hat{\bm X}_{\ell}(-v) {\hat{\bm D}}_{\ell}$ to obtain the estimate for $\bm x_{\ell}(v)$, i.e.,
\begin{equation}
\label{eq:two-step2}
    \hat{\bm x}_{\ell}^{(t)}(v) =  {\hat{\bm D}_{\ell}}^{-1}(\hat{{\bm X}}_{\ell}(-v)'\hat{{\bm X}}_{\ell}(-v))^{-1}\hat{\bm X}_{\ell}(-v)'\hat{\bm b}_{\ell\{v\}}.
\end{equation}

The optimization in (\ref{eq:two-step1})  obtains an intermediate estimate $\hat{\bm b}_{\ell\{v\}}$ which is an estimate of the $\ell$th latent source signal which satisfies the desired element-wise sparsity but does not have the low-rank structure. Then, in (\ref{eq:two-step2}), we project the intermediate estimate onto the low-rank space to obtain an updated latent subspace coordinate estimate for the $v$th node in the $\ell$th source signal.  

After updating $\bm x_{\ell}(v)$, we rotate to the next node and repeat the procedure described above across nodes $v=1,\ldots,V$ to obtain updated estimate for $\bm X_{\ell}$. An advantage of the proposed node-rotation algorithm is that it has analytic solutions and does not need gradient-based numerical approximation, which makes it highly efficient and reliable. 

{\bf Step 2}: {\it Updating $\bm D_{\ell}$}. The second step is to update the diagonal matrix $\bm D_{\ell}$ for $\ell = 1,...,q$, given the estimate of $\bm X_{\ell}$ from the $t$th iteration and the estimate of $\widetilde{\bm A}$ from the $t-1$ iteration. We update the estimate of the diagonal of $\bm D_{\ell}$, i.e. $\bm d_{\ell} = \text{diag}(\bm D_{\ell})$ via the following, 
\begin{align} \label{update5}
     \min\limits_{\bm d_{\ell} \in \mathcal{R}^{R_{\ell}}} \Big\|   \widetilde{\bm Y}'\hat{\widetilde{\bm a}_{\ell}} - \hat{\bm Z}_{\ell} \bm d_{\ell} \Big\|_2^2 + \phi \| \hat{\bm Z}_{\ell}\bm d_{\ell} \|_1,
\end{align}
where $\hat{\bm Z}_{\ell} \in \mathcal{R}^{p \times R_{\ell}}$ with the $r$th column of $\hat{\bm Z}_{\ell}$ being $\L(\hat{\bm x}_{\ell}^{(r)} \hat{\bm x}_{\ell}^{(r)'})$ $(r=1,\ldots,R_{\ell})$. A similar procedure for solving (\ref{update2}) can be adopted for solving (\ref{update5}). 

{\bf Step 3}: {\it Updating $\widetilde{\bm A}$}. This step is to update mixing matrix $\widetilde{\bm A}$ given the estimates of $\bm X_{\ell}$ and $\bm D_{\ell}$ from the $t$th iteration. Specifically, \begin{equation}
\label{eq:est_mix}
    \hat{\widetilde{\bm A}} = \widetilde{\bY} \hat{\bm S}^{(t)'}  {(\hat{\bm S}^{(t)}\hat{\bm S}^{(t)'})}^{-1},
\end{equation}
where $\hat{\bm S}^{(t)} = [\hat{\bm s}^{(t)}_{1}, \ldots, \hat{\bm s}^{(t)}_{q}]'$ with $\hat{\bm s}_{\ell}=\L(\hat{\bm X}_\ell^{(t)}\hat{\bm D}_\ell^{(t)}\hat{\bm X}_{\ell}^{(t)'})$.
The solution from (\ref{eq:est_mix}) is then orthogonalized to obtain the updated mixing matrix. 

\begin{algorithm}
   \caption{An Iterative Node-Rotation Algorithm for Learning LOCUS}
   \label{locus_alg}
\begin{algorithmic}
    \STATE {\bfseries Initial}: Initialize $\hat{\widetilde{\bm A}}^{(0)}, \{\hat{\bm X}_{\ell}^{(0)}, \hat{\bm D}_{\ell}^{(0)}\}$ based on estimates from existing methods such as connICA.
   \REPEAT 
     \STATE For $\ell = 1...q$, 
     \STATE \hspace{0.2in} For $v=1,\ldots, V$,\\
       {\bf Step 1. Update } $\bm x_{\ell}(v)$ : \\
       $$\hat{\bm b}^{(t)}_{\ell\{v\}}=\argmin\limits_{\bm b \in \mathcal{R}^{V-1}} \Big\|\widetilde{\bm Y}_{\{v\}}'\widetilde{\bm a}_{\ell}   -  \bm b \Big\|_2^2 + \phi \Big\|\bm b \Big\|_1,$$
        $$  \hat{\bm x}^{(t)}_{\ell}(v) =  {\hat{\bm D}_{\ell}}^{-1}(\hat{{\bm X}}_{\ell}(-v)'\hat{{\bm X}}_{\ell}(-v))^{-1}\hat{\bm X}_{\ell}(-v)'\hat{\bm b}^{(t)}_{\ell\{v\}}.$$
        \\
      \hspace{0.2in} End for $v$
     \STATE    {\bf Step 2. Update} ${\bm D}_{\ell}$:\\
    $$ \hat{\bm d}^{(t)}_{\ell}=\textrm{diag}(\hat{\bm D}^{(t)}_{\ell})= \argmin\limits_{\bm d_{\ell} \in \mathcal{R}^{R_{\ell}}} \Big\|   \widetilde{\bm Y}'\hat{\widetilde{\bm a}_{\ell}} - \hat{\bm Z}_{\ell} \bm d_{\ell} \Big\|_2^2 + \phi \| \hat{\bm Z}_{\ell}\bm d_{\ell} \|_1,$$
    \\
   End for $\ell$
   
     \STATE \textbf{Step 3. Update} $\widetilde{\bm A}$: 
      \STATE  
      \begin{equation*}
    \hat{\widetilde{\bm A}} = \widetilde{\bY} \hat{\bm S}^{(t)'}  {(\hat{\bm S}^{(t)}\hat{\bm S}^{(t)'})}^{-1},
      \end{equation*}
       $$ \textrm{where} \,\, \hat{\bm S}^{(t)} = [\hat{\bm s}^{(t)}_{1}, \ldots, \hat{\bm s}^{(t)}_{q}]' \quad \textrm{with} \,\,\hat{\bm s}^{(t)}_{\ell}=\L(\hat{\bm X}_\ell^{(t)}\hat{\bm D}_\ell^{(t)}\hat{\bm X}_{\ell}^{(t)'})$$.
       
     \hspace{0.3in} Perform an orthogonal transformation on $\hat{\widetilde{\bm A}}^{(t)}$
      
\UNTIL{$\frac{\|\hat{\widetilde{\bm A}} ^{(t)}-\hat{\widetilde{\bm A}}^{(t-1)}\|_F}{\|\hat{\widetilde{\bm A}}^{(t-1)}\|_F} <\epsilon_1$ and $\frac{\|\hat{\bm S}^{(t)}-\hat{\bm S}^{(t-1)}\|_F}{\|\hat{\bm S}^{(t-1)}\|_F} <\epsilon_2$}
\vspace{0.1 cm}
\end{algorithmic}
\end{algorithm}

The proposed iterative estimation algorithm is summarized in Algorithm \ref{locus_alg}.  The LOCUS learning in (\ref{eq:locus_preprocess}) is non-convex optimization problem. To solve it, we propose the estimation algorithm based on the block multi-convex structure of the optimization function. A function is block multi-convex if it is convex with respect to each of the individual arguments while holding all others fixed. The formal definition of block multi-convexity  \citep{gorski2007biconvex} is provided below. 

\begin{propdef}[Block Multi-Convexity] \label{biconvex_def}
Define a partition of set $\bm x = \{x_1,...,x_p\} \in \mathcal{R}^p$ as a collection of disjoint non-empty subsets $\{\bm x_1,..,\bm x_h\}$ of $\bm x$ with 1): $\bigcup_{i=1}^h \bm x_i =  \bm x$; 2): $\bm x_i \subseteq \bm x$ and $\bm x_i\ne \emptyset $; 3) $\bm x_i \cup \bm x_j = \emptyset$. A function $f(x_1,...,x_p) \colon \mathcal{R}^p \mapsto \mathcal{R}$ is a block multi-convex function if there exists a {\bf partition} $\{\bm x_1,..,\bm x_h\}$ on $\{x_1,...,x_p\}$ satisfying that $f$ is convex with respect to each of the individual $\bm x_i$, $i = 1,...,p$, while holding all others fixed.
\end{propdef}

In Proposition \ref{Theo:biconvex}, we show that the LOCUS optimization function has the block multi-convexity structure.   
\begin{prop} \label{Theo:biconvex}
Let $f(\widetilde{\bm A},\{\bm X_\ell,\bm D_\ell \})$ be the objective function in (\ref{eq:locus_preprocess}). The function $f$ is block multi-convex with respect to the partition of $\mathcal{P} = \{ \bm x_1(1),..,\bm x_1(V),\ldots,\bm x_q(V), \bm d_1,..,\bm d_q, \widetilde{\bm A} \}$. 
\end{prop}

The proof of Proposition \ref{Theo:biconvex} is presented in the Appendix. Our proposed node-rotation iterative estimation algorithm exploits the block multi-convexity of the objective function to solve the non-convex optimization problem for learning LOCUS. 

\subsubsection{Tuning Parameter Selection}
In LOCUS, the rank parameters $\{R_{\ell}\}_{\ell=1}^{q}$ control the dimension of the reduced subspace of the low-rank structure for the latent connectivity sources. Given the difference in the topology and structure across the neurocircuitry traits, it is oversimplified to specify a common rank parameter for all connectivity sources. Therefore, we propose an adaptive selection approach to choose $\{R_{\ell}\}$ for the $q$ latent sources. Specifically, when updating $\hat{\bm s}_\ell \,(\ell=1,\ldots,q)$, we denote $\hat{\bm s}_\ell^{\ast}$ to be the latent sources estimated without the low-rank structure assumption, which can be obtained from the intermediate source estimates in (\ref{eq:two-step1}) which have the desired element-wise sparsity but do not have the low-rank structure. The rank $R_{\ell}$ is chosen to achieve a desired level of closeness between the unstructured $\hat{\bm s}^{\ast}$ and the latent source with the low-rank structure, i.e. $\hat{\bm s}_{\ell}=\L(\hat{\bm X}_{\ell} \hat{\bm D}_{\ell}\hat{\bm X}_{\ell}')$. Specifically, $R_{\ell}$ is selected to be the smallest integer value such that,
\begin{equation}
\label{rank_sel}
   \| \hat{\bm s}_{\ell} - \hat{\bm s}^{\ast}_{\ell} \|_2^2 / \|\hat{\bm s}^{\ast}_{\ell} \|_2^2 \le 1 - \rho,
\end{equation}
where $\rho \in (0,1)$ is a proportion parameter controlling the desired level of closeness between the unstructured and low-rank structured latent sources. Once the proportion parameter $\rho$ is specified, the proposed approach adaptively selects the rank for each of the latent sources. The proposed adaptive selection method not only allows LOCUS to flexibly capture connectivity traits with different topology characteristics but also simplifies the challenging task of selecting $q$ rank parameters $\{R_{\ell}\}_{\ell=1}^{q}$ to only selecting a single parameter $\rho$.

With the proposed approach, the tuning parameters for learning the LOCUS model include $\phi$ and $\rho$. We
propose to select those parameters via a BIC-type criterion that balances between model fitting and model sparsity,
\begin{equation}
    \textrm{BIC}= -2 \sum_{i=1}^N \text{log} \Big(g(\bm y_i;\sum_{\ell=1}^{q}\hat{a}_{i\ell}\hat{\bm s}_{\ell}, \hat{\sigma}^2 \bm I_p )\Big) + \text{log}(N) \sum_{\ell=1}^{q}\|\hat{\bm s}_{\ell}\|_0 
\end{equation}
where $g$ denotes the pdf of a multivariate Gaussian distribution, $\hat{\sigma}^2 = \frac{1}{Np}\sum_i \|\bm y_i-\sum_{\ell=1}^{q}\hat{a}_{i\ell}\hat{\bm s}_{\ell}\|_2^2$, $\|\cdot\|_0$ denotes the $L_0$ norm . This criterion balances between model fitting and model sparsity. Similar criteria have been employed in the tuning parameter selection on sparse tensor decomposition \citep{allen2012sparse,kim2013sparse,sun2017store}.

\section{Application to rs-fMRI connectivity data from the Philadelphia Neurodevelopmental Cohort (PNC)}\label{realsec}

We applied the proposed method to resting state fMRI (rs-fMRI) data collected in the Philadelphia Neurodevelopmental Cohort (PNC) study. 

\subsection{PNC Study and Data Description}
The PNC is a collaborative project from the Brain Behavior Laboratory at the University of Pennsylvania and the Children's Hospital of Philadelphia (CHOP), funded by NIMH through the American Recovery and Reinvestment Act of 2009 \citep{satterthwaite2014neuroimaging, satterthwaite2014linked}. The PNC study includes a population-based sample of individuals aged 8–21 years selected among those who received medical care at the Children's Hospital of Philadelphia network in the greater Philadelphia area; the sample is stratified by sex, age and ethnicity. A subset of participants from the PNC were recruited for a multimodality neuroimaging study which included resting-state fMRI (rs-fMRI). 

Prior to analysis, we performed quality control on the rs-fMRI including displacement analysis to remove images with excessive motion \citep{satterthwaite2014linked,wang2016efficient}. Among the subjects who had rs-fMRI scans, 514 participants' data met our quality control criterion and were used in the following analysis. Among these subjects, 289 (56\%) were female and the mean age was 15.3 years (SD = 3.1).

The rs-fMRI data were processed using standard preprocessing procedure. Specifically, skull stripping was performed on the T1 images to remove extra-cranial material, then the first four volumes of the functional time series were removed to adjust for initial stabilization, leaving 120 volumes for subsequent preprocessing. The anatomical image was registered to the 8th volume of the functional image and subsequently spatially normalized to the MNI standard brain space. These normalization parameters from MNI space were used for the functional images, which were smoothed with a 6 mm FWHM Gaussian kernel. Motion corrections were applied on the functional images. A validated confound regression procedure \citep{satterthwaite2015linked} was performed on each subject's time series data to remove confounding factors including motions, global effects, white matter (WM) and cerebrospinal fluid (CSF) nuisance signals. Furthermore, motion-related spike regressors were included to bound the observed displacement. Lastly, the functional time series data were band-pass filtered to retain frequencies between 0.01 and 0.1 Hz which is the relevant frequency range for rs-fMRI.

In our paper, we adopt Power's 264-node brain parcellation system \citep{power2011functional} for connectivity analysis. Each node is a 10 mm diameter sphere in the standard MNI space representing a putative functional area, and the collection of nodes provides good coverage of the whole brain. The nodes are assigned into 10 functional modules that correspond to the major resting state networks \citep{smith2009correspondence}. The functional modules include medial visual network (“Med Vis”), occipital pole visual network (“OP Vis”), lateral visual network (“Lat Vis”), default mode network (“DMN”), cerebellum (“CB”), sensorimotor network (“SM”), auditory network (“Aud”), executive control network (“EC”), and right and left frontoparietal networks (“FPR” and “FPL”). 232 of the 264 nodes that are associated with the resting state networks were used for our connectivity analysis. We extract the fMRI time series from each node and obtain $232 \times 232$ connectivity matrix for each subject by evaluating the  pair-wise correlations between the node-specific fMRI series. Fisher's Z transformation is applied to the correlations to obtain the connectivity data for LOCUS decomposition.

% Reproducibility
\begin{figure}[h!]
\begin{center}
 \includegraphics[scale=0.2 ]{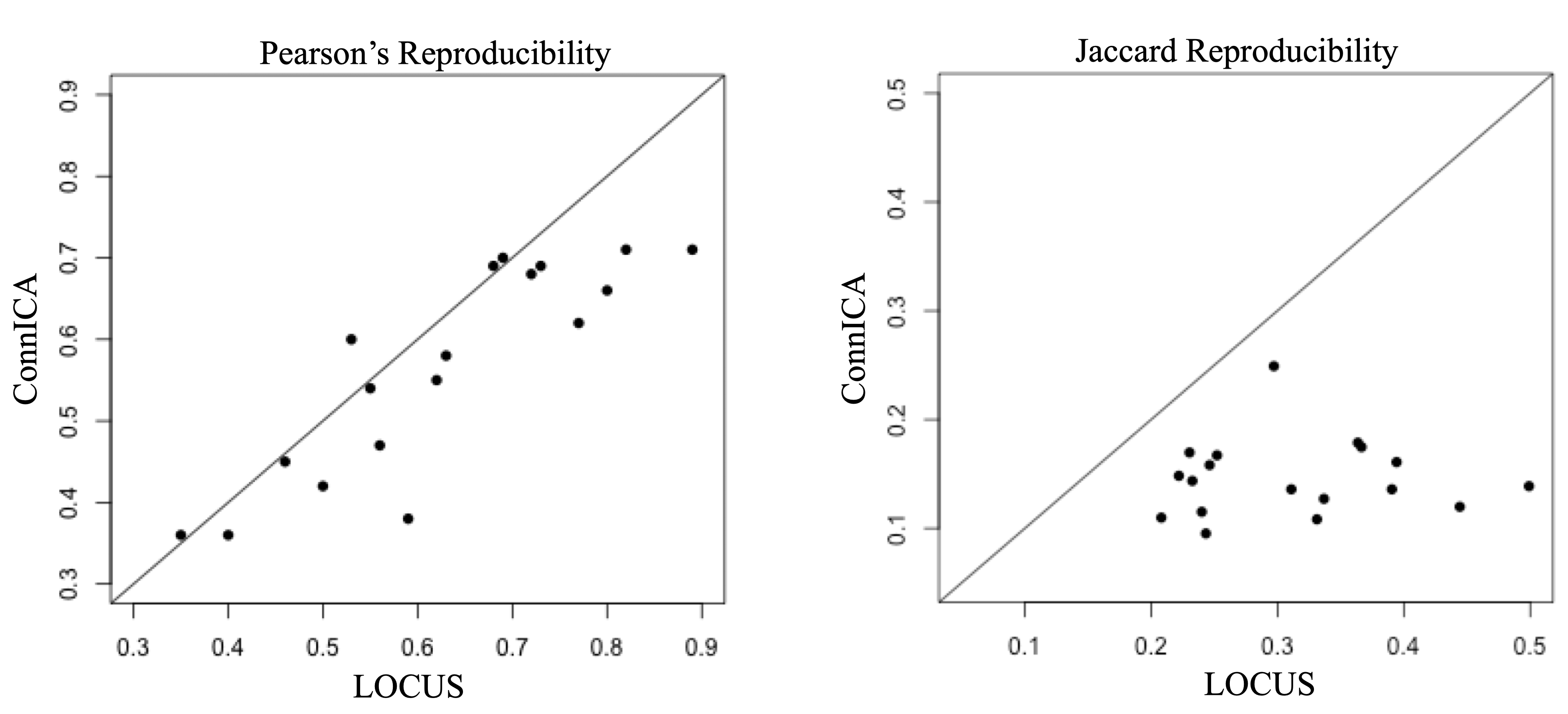}
    \caption{Reproducibility analysis for 18 matched latent sources from LOCUS and connICA. It is shown that for the matched latent sources LOCUS generally has significant higher reproducibility compareed to connICA approach (p-value: 0.005 for Pearson correlation, $< 0.001$ for Jaccard Index). }
    \label{Fig:real_reprodu}
\end{center}
\end{figure}

\subsection{Connectivity Analysis for PNC Study}

We apply the proposed LOCUS method to decompose the preprocessed functional connectivity data from PNC study. We also implement connICA which represents the currently leading method in neuroimaging for decomposing brain connectivity data. For the LOCUS model, we select tuning parameter $\phi=2$ and $\rho=0.85$ based on the proposed BIC criterion. The number of latent sources is selected as $q=30$ based on the reproducibility and interpretability of the extracted sources. The source-specific rank parameters $\{R_{\ell}\}$ are determined using the proposed adaptive selection approach. Across the latent sources, the mean (SD) of $R_\ell$ is $8.7 (1.6)$ where the min, median and max being 6, 8 and 12, respectively. We evaluate the reproducibility of the 30 latent sources estimated by LOCUS and connICA where the reproducibility of each extracted latent sources is assessed using a reliability index proposed by \citet{kemmer2018evaluating}. The index provides a scaled and chance-corrected measure to assess the reproducibility of latent sources extracted with blind source separation methods. Specifically, the reliability index of the $\ell$th latent source estimates is evaluated based on $B=200$ replications which correspond to bootstrap samples in the real data analysis: 
\begin{equation}
\label{eq:reproducibility}
RI_{\ell} = \frac{ \frac{1}{B}\sum_{b=1}^B \{ h(\bm S_{\ell}, \hat{\bm S}^{(b)}_{\ell}) \} -  \frac{1}{Bq}\sum_{b=1}^B\sum_{j=1}^q h(\bm S_{\ell}, \hat{\bm S}^{(b)}_{j})} {1-\frac{1}{Bq}\sum_{b=1}^B\sum_{j=1}^q h(\bm S_{\ell}, \hat{\bm S}^{(b)}_{j})}, 
\end{equation}
here $\bm S_{\ell}$ is the estimated latent source extracted by a blind source separation method (i.e. LOCUS or connICA) from the original data, and $\hat{\bm S}^{(b)}_{\ell}$ is the latent source estimated by the method from the $b$th replicate (i.e. the bootstrap data sample) that is matched with the original $\bm S_{\ell}$. We use a greedy matching algorithm to match latent sources \citep{Keeratimahat2021discussion, wu2021distributionalrejoinder}. Specifically, we find the pair of a bootstrap latent source and an original latent source that has the highest correlation coefficient, then remove the best matched pair from the following matching process, and repeat until all the latent source pairs are matched. In equation (\ref{eq:reproducibility}), $h(\cdot)$ is a measure of similarity between two sources, which can be specified as Pearson correlation \citep{wang2016efficient} or Jaccard Index \citep{real1996probabilistic}. The reliability index reflects the similarity between the two source signals, removing by-chance similarity between the original source and any source estimates out of the $q$ extracted latent sources. It is further scaled by its maximum possible value so that it typically ranges from 0 to 1, where $RI_\ell$ = 0 indicates the $\ell$th latent source is not reproducible across the replications after we correct for by-chance correlations and $RI_{\ell}$ close to 1 indicates that the component is highly reproducible across replication data. For comparison purpose, we match the estimated latent sources from LOCUS and connICA based on their correlations and identify 18 matching sources with correlations of 0.6 and above.

Figure \ref{Fig:real_reprodu} compares the reproducibility of the 18 matched latent sources extracted by LOCUS and connICA using reliability indices based on both Pearson correlation and Jaccard Index. The connectivity traits from LOCUS show significantly higher reproducibility compared with those from connICA  (p-value: $0.005$ for Pearson correlation, $<0.001$ for Jaccard Index). For most of the 18 traits, the LOCUS's estimates demonstrate about 15\% to 50\% increase in reliability as compared with those from connICA. The complete reproducibility results for all the 30 latent sources extracted by LOCUS and connICA can be found in the Appendix.

Figure \ref{fig:matchedIC} displays six highly reproducible latent sources by LOCUS, which have a reliability index of 0.7 or higher. For comparison purposes, we also display the matched latent sources extracted by connICA. Compared to the noisy estimates by connICA, the latent sources,i.e. connectivity traits, extracted by LOCUS is much sparser and more clearly reveal the key neural connections contributing to each of the connectivity trait. In Figure \ref{fig:matchedICview}, we visualize the top 1\% edges in the six highly reproducible connectivity traits extracted by LOCUS using BrainNet Viewer \citep{xia2013brainnet}. The connectivity traits reflect various neural circuits underlying the observed overall brain connectivity. Each neural circuit represents a set of brain connections that tend to occur together including both within module connections as well as between module connections. Trait 1 mainly consists of connections involving nodes from the FPR and Aud functional modules. Trait 2 mainly consists of connections within the Med Vis module and connections between Med Vis and some brain regions including the other visual modules, i.e. Op Vis and Lat Vis., and medial frontal cortex regions in the EC module. Trait 3 is a cerebellum-related connectivity trait including connections within CB and also between CB and the other brain regions. Trait 4 consists of the connections between Op Vis and other brain regions. Trait 5 mainly consists of edges connecting Med Vis with sensory and motor modules including SM, CB, Aud and also with anterior cingulate regions in EC. Trait 6 mainly consists of connections among the cognitive functional modules including EC, DMN, FPL and FPR. These connectivity traits are of great interest to understand the brain network separation. 

% 6 Matched ICs - Heatmap with connICA
\begin{figure}
\begin{center}
\includegraphics[scale=0.5]{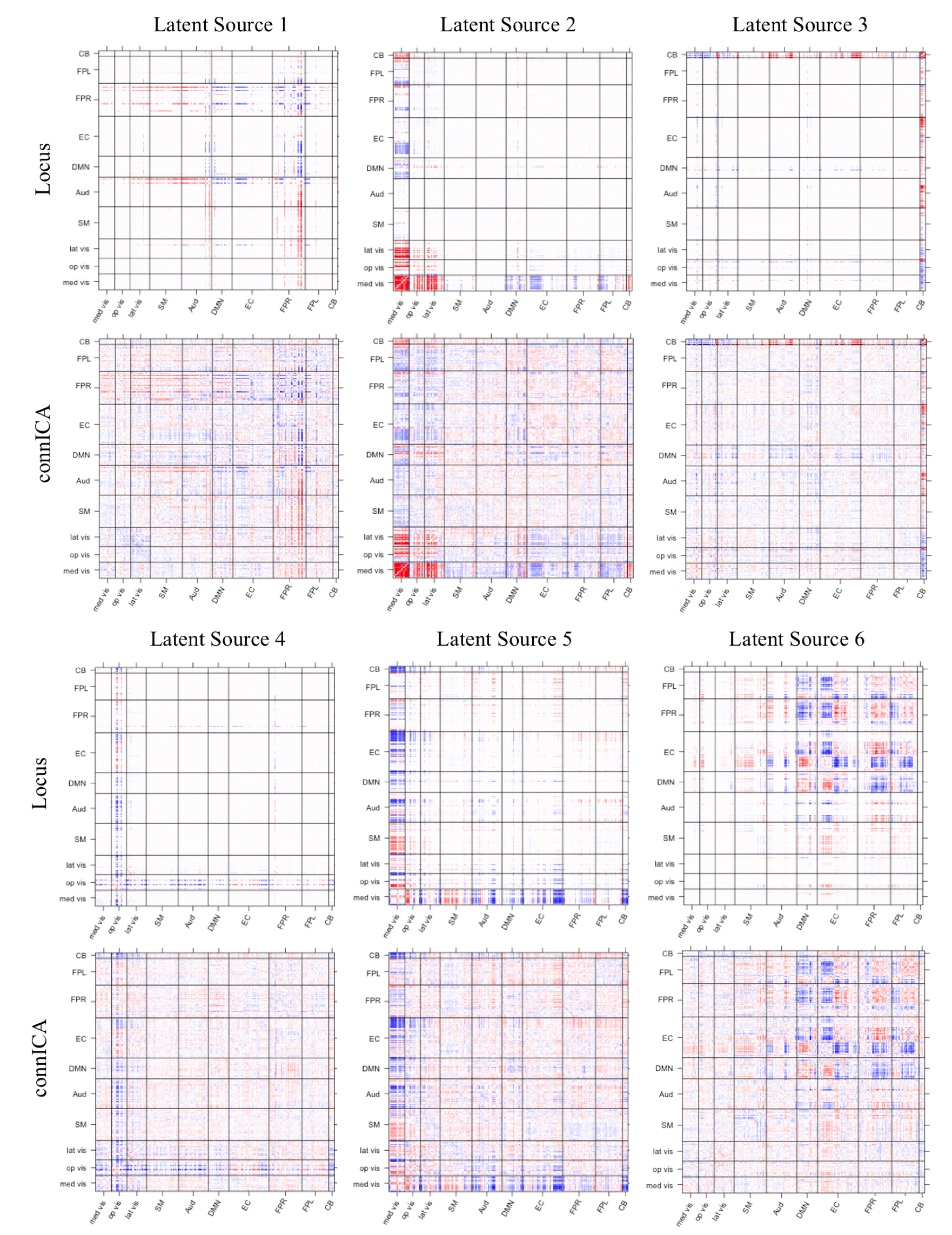}
\caption{\small{Six matched latent sources, i.e. connectivity traits,  estimated by LOCUS and connICA with high reproducibility. The six connectivity traits estimated by LOCUS have a reproducibility higher than 0.7. }}\label{fig:matchedIC}
\end{center}
\end{figure}

Importantly, LOCUS derived subjects' loadings for latent source 3, i.e. the cerebellum-related connectivity trait, are found to be significantly associated with subjects' age ($p = 0.027, \text{FDR adjusted }p = 0.062$). Our find coincide with previous results in the neuroscience literature that show age-related changes concentrated in cerebellum during adolescence \citep{kundu2018integration}. However, previous results from \citet{kundu2018integration} were based on standard ICA decomposition of fMRI BOLD  series, whereas our findings are derived from unsupervised brain connectome decomposition. Moreover, the cerebellum connectivity trait recovered by our LOCUS method reveals the connection between cerebellum  and other brain functional modules including EC, Aud and Visual, which are not discovered in the previous study. It is worth noting that subjects' loadings on the corresponding connectivity trait from connICA fail to reveal such association ($p = 0.952, \text{FDR adjusted }p = 0.982$). The results show that with the advantages provided by the low-rank structure and the novel sparsity regularization, the proposed LOCUS method generates more accurate and parsimonious results in recovering underlying connectivity traits. This not only reduces false positive findings in brain connections but may also lead to more precise estimates of subject trait loadings that help reveal the linkage between brain connectivity traits and demographic and clinical factors.

Moreover, LOCUS also identifies several reproducible connectivity traits that are not revealed by connICA. Figure \ref{fig:unmatchedIC} shows two latent sources (LS) estimated from LOCUS which have relatively high reproducibility, with reliability of 0.64 (LS7) and 0.55 (LS8), but are not identified by connICA. Figure \ref{fig:unmatchedICview} displays the top 1\% edges of these two latent sources. LS7 contains connections involved with the visual networks including connections within Lat Vis and anterior-posterior connections between Lat Vis and EC, FPR and DMN. Analysis of the subjects' loadings on LS7 shows that this connectivity trait is significantly associated with subject's gender ($p = 0.005, \text{FDR adjusted }p = 0.002$). A similar finding related to gender differences in the visual network has been discussed in \citet{ingalhalikar2014sex} based on structural connectome data. Our LOCUS results provide new evidence from the functional connectome and also generate new findings revealing potential gender differences in the connections between the visual network and other modules (EC, FPR and DMN).  LS8 mainly contains connections involving the EC network including connections within EC and the connections between EC and SM, Op Vis, Aud. The subjects' loadings of LS8 are significantly associated with their age  ($p < 0.0001, \text{FDR adjusted }p = 0.0001$), indicating neurodevelopment in EC connections.

% 6 Matched ICs - BrainNetViewer LOCUS 1%
\begin{figure}
\begin{center}
 \includegraphics[scale=0.38]{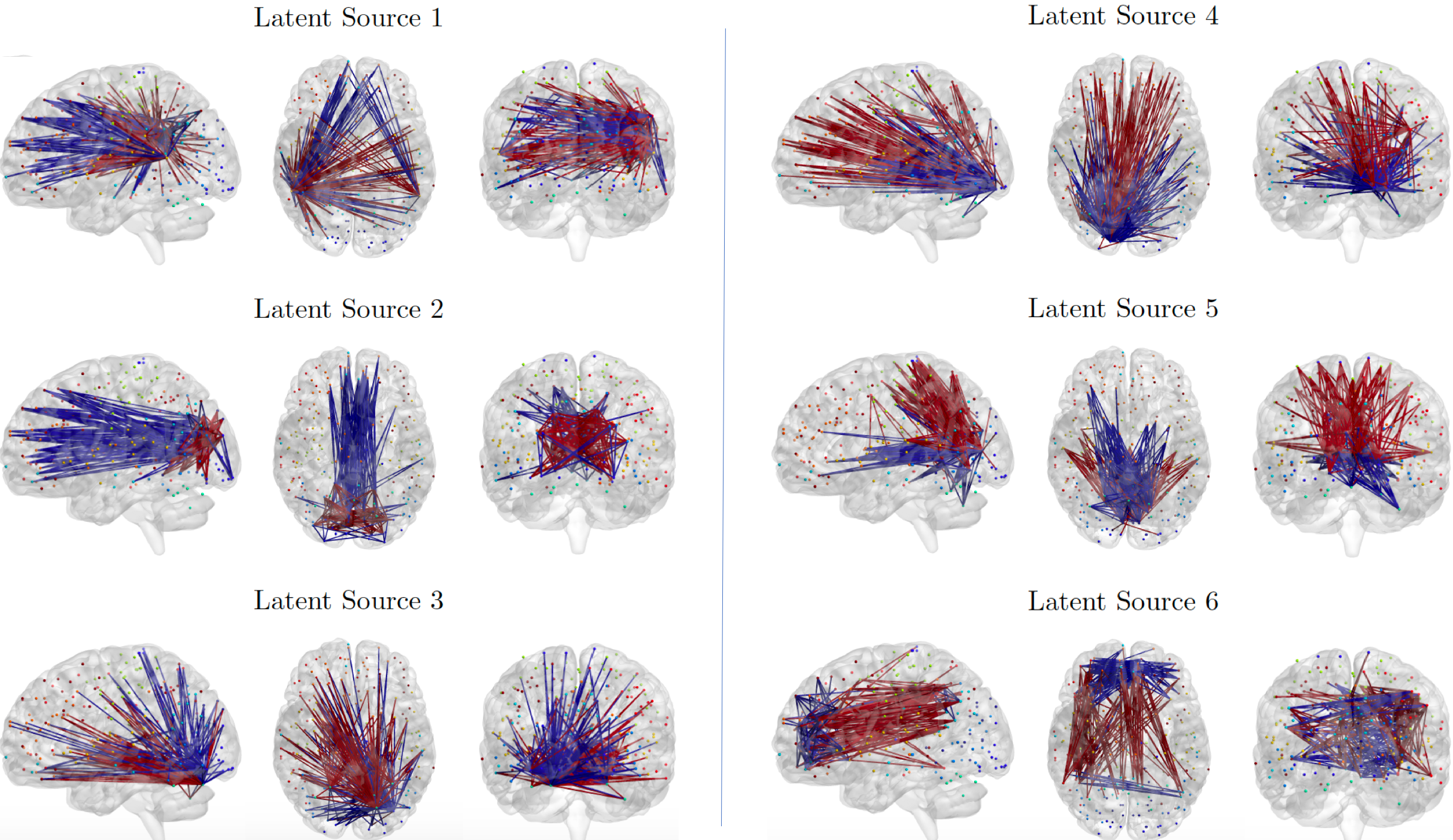}
\caption{\small{The top 1\% brain connections of the 6 highly reproducible latent sources, i.e. connectivity traits, based on LOCUS (blue are negative connections, red are positive connections).}}\label{fig:matchedICview}
\end{center}
\end{figure}

\begin{figure}
\begin{center}
\includegraphics[scale=0.45]{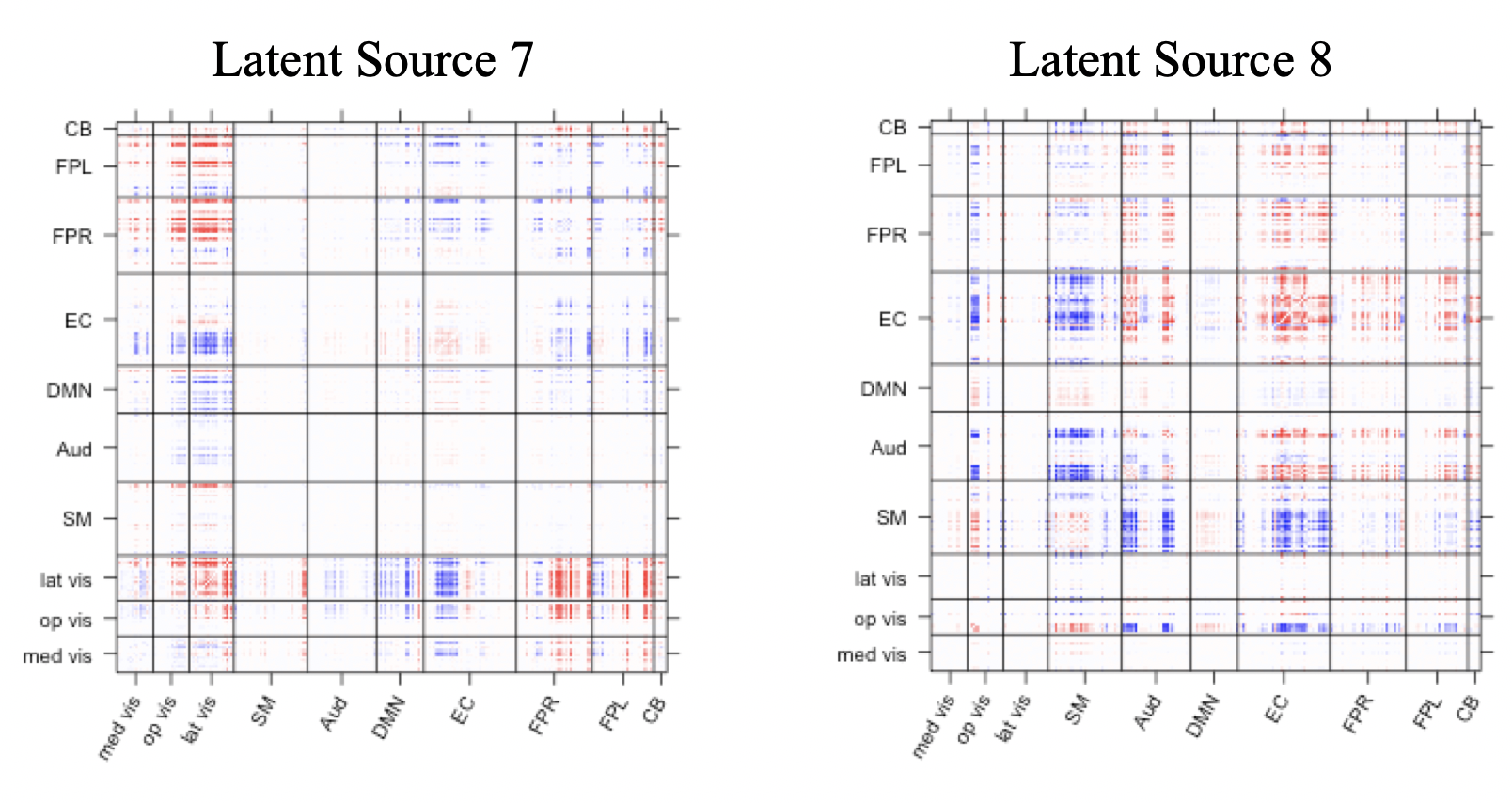} 
\caption{\small{Two LOCUS extracted latent sources, i.e. connectivity traits, that are not identified by connICA. These two latent sources have relatively high reproducibility and are significantly associated with subjects' gender or age.  }}\label{fig:unmatchedIC}
\end{center}
\end{figure}

\begin{figure}
\begin{center}
\small
\includegraphics[scale=0.5]{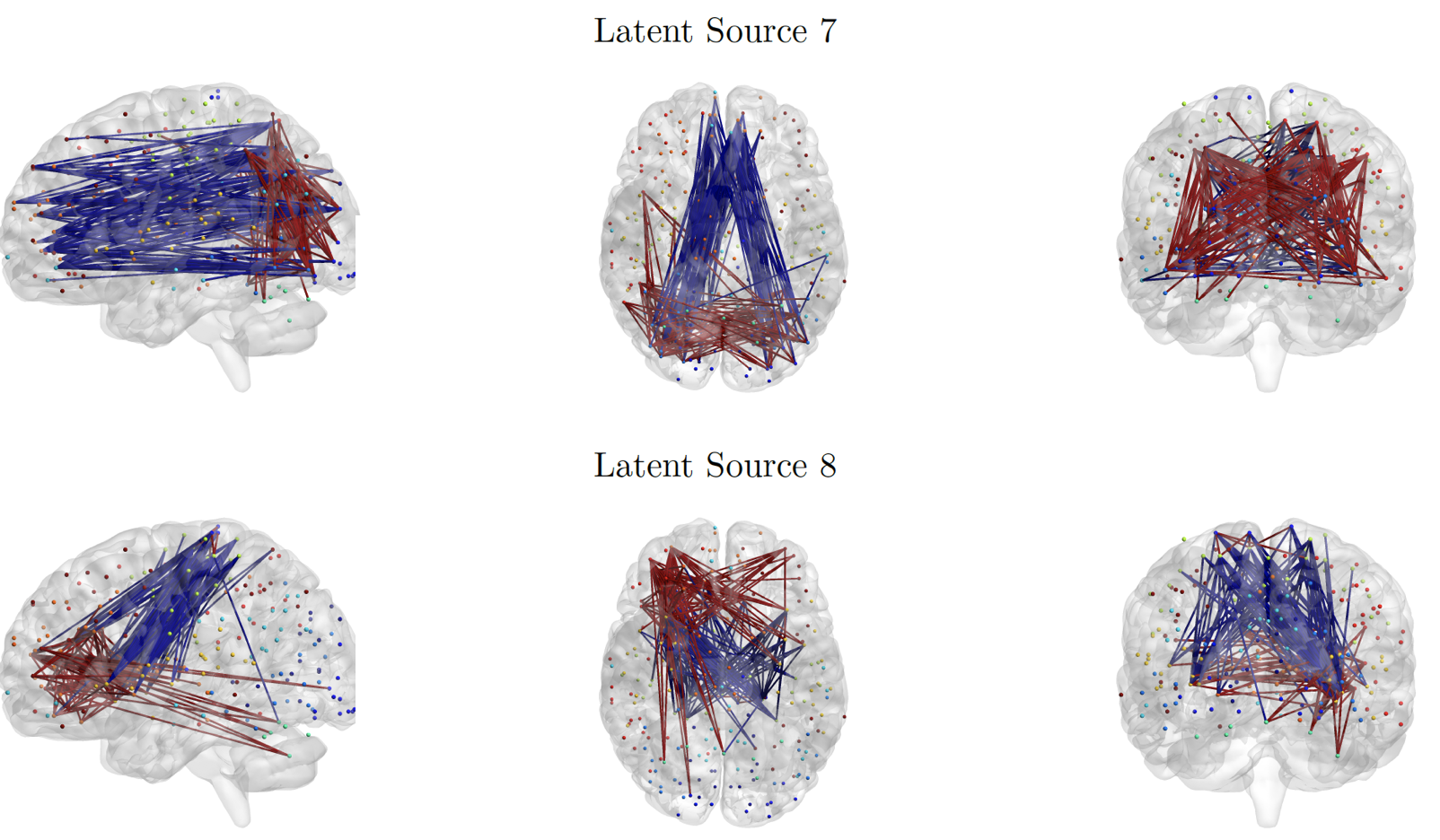}
\caption{\small{Visualization of the top 1\% brain connections of the two LOCUS extracted latent sources that are not identified by connICA (blue are negative connections, red are positive connections).}}\label{fig:unmatchedICview}
\end{center}
\end{figure}

\section{Simulation Study}\label{sec3}
In this section, we investigate the performance of our model based on simulation studies. We compare the performance of LOCUS with two other source separation methods: connICA and the dictionary learning (``DL") method \citep{mairal2009online} which is a popular sparse decomposition method. Moreover, to evaluate the performance of the proposed angle-based sparsity regularization, we compare LOCUS with low-rank decomposition of brain connectivity with the two existing sparsity penalization, i.e. vector-wise L1 penalty (``LOC-VecL1") and the nuclear norm penalty (``LOC-Nuclear"). Specifically, the vector-wise L1 regularization aims to achieve sparsity in the vectors in the low-rank structure and the penalizing term is based on $\sum_{\ell=1}^q\|\bm X_{\ell}\|_1$. The nuclear norm penalization aims to achieve low-rankness in the source signals and the penalizing term is based on $\sum_{\ell=1}^q\|\bm S_{\ell}\|_*$. As with LOCUS, the BIC criteria was used to select the tuning parameters for these existing sparsity regularization. 

\begin{figure}
\begin{center}
\small
\includegraphics[scale=0.5]{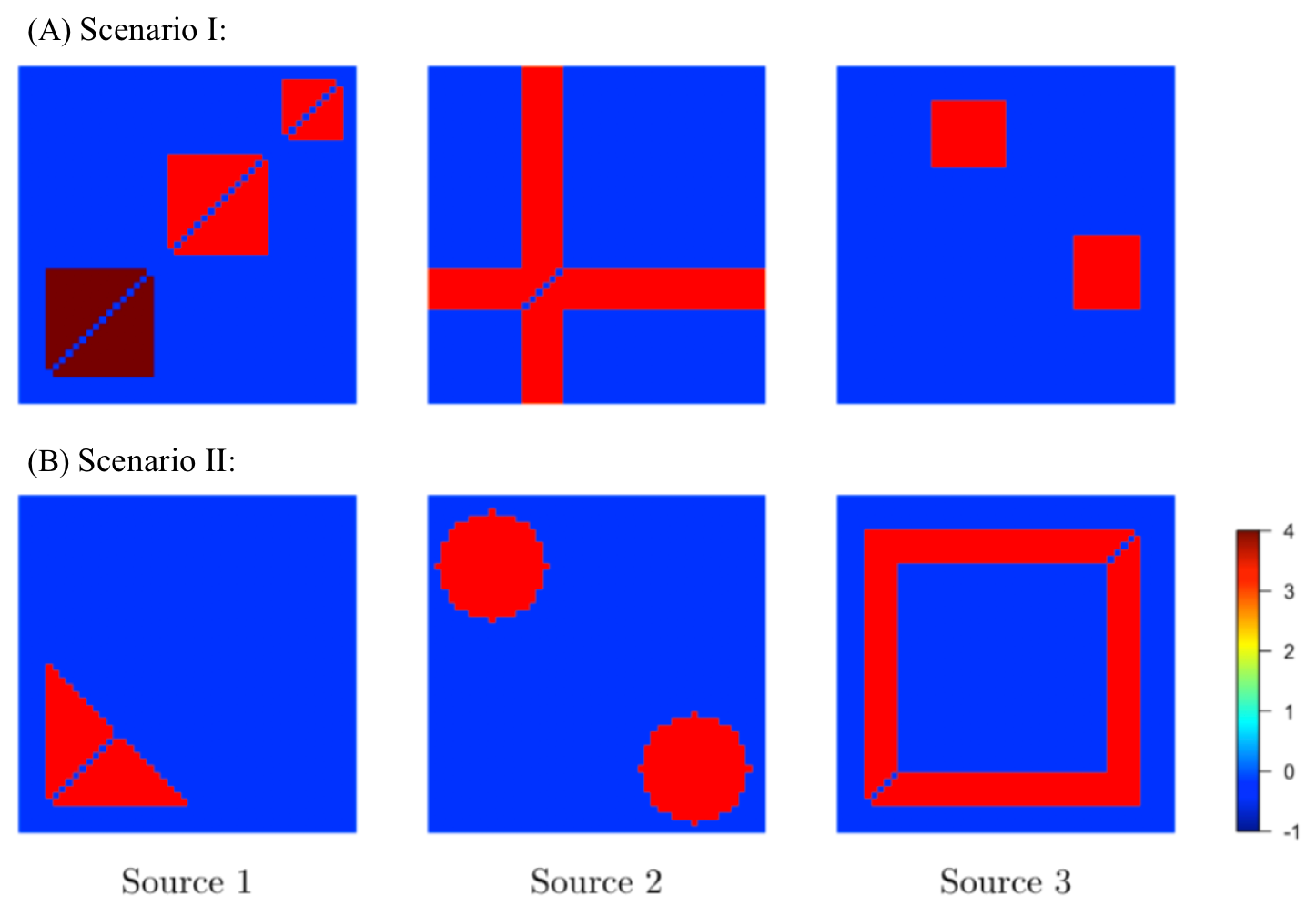}
\caption{\small{True underlying  source signals for the two simulation scenarios. }}\label{fig:simusetting}
\end{center}
\end{figure}

\subsection{Synthetic Data}
We specified $V = 50$, $q = 3$ and two sample sizes, $N = 50, 100$. We considered two simulation scenarios in terms of the underlying source signals. In scenario I, motivated by connectivity traits extracted from real imaging data (Figure \ref{fig1:illustrate2}), we generated three latent connectivity source signals with the diagonal block shape, crossing shape and off-diagonal block shape as shown in Figure \ref{fig:simusetting} (A). These patterns are commonly observed in neuroimaging connectivity \citep{amico2017mapping, amico2018mapping} and hence serve as a realistic scenario for brain imaging applications. In scenario II, we considered three latent source signals with diagonal triangle shape, off-diagonal circle shape and long-range hollow square shape shown in Figure \ref{fig:simusetting} (B). These shapes were shown to be several particularly challenging patterns to be captured by the low-rank structure  \citep{zhou2013tensor}. Therefore, scenario II allows us to evaluate the robustness of the performance of the proposed LOCUS model under challenging scenarios that deviate from the low-rank assumption. The mixing coefficients were also sampled from estimates from real imaging data. Furthermore, we added zero mean Gaussian noises to the mixture of signals where the variance was specified based on signal-to-noise ratio observed from real data. Specifically, we considered three variance settings with $\sigma^{2}=1,3^2$ and $6^2$, corresponding to low, medium and high variance levels, respectively. In summary, we have $2 \times 2 \times 3=12$ simulation settings with various combinations of sample sizes, source signal patterns and variance levels and for each setting we generated 100 simulation runs to assess the variations in performance. In addition, we considered simulation settings where there are variations in the signal intensity across
edges in the latent sources. Please refer to the Appendix for the additional simulation studies.

% all results for set 1
\begin{table}
	\caption{Simulation results for comparing LOCUS with other methods with 100 simulation runs for Scenario I. Values presented are mean and standard deviation of correlations between the true and estimated latent sources and loading/mixing matrices. }
	\label{tbl:set1}
	\centering
		\includegraphics[scale=0.52]{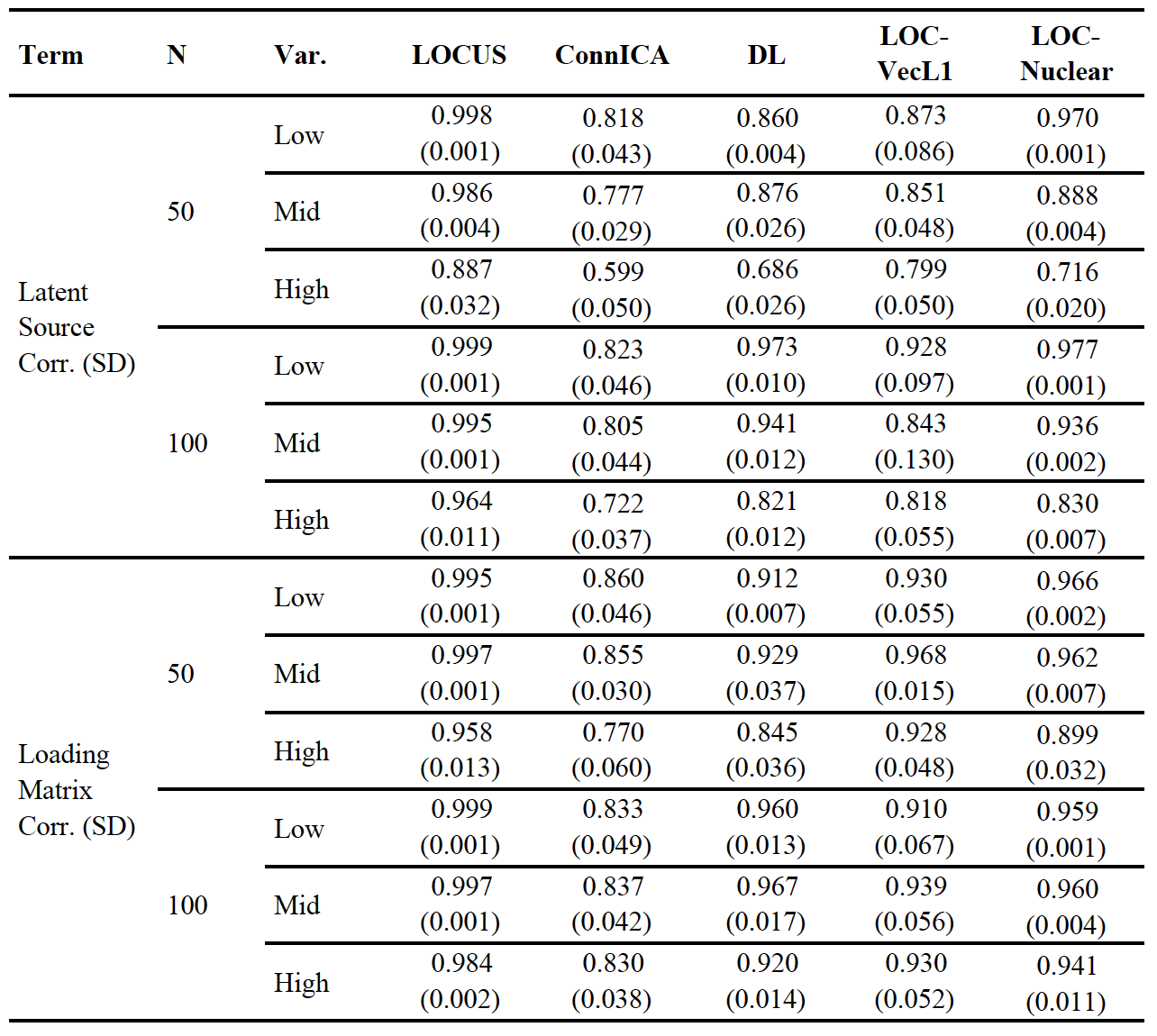}
\end{table}

% all results for set 2
\begin{table}
	\caption{Simulation results for comparing LOCUS with other methods with 100 simulation runs for Scenario II. Values presented are mean and standard deviation of correlations between the true and estimated latent sources and loading/mixing matrices. }
	\label{tbl:set2}
	\centering
	\includegraphics[scale=0.52]{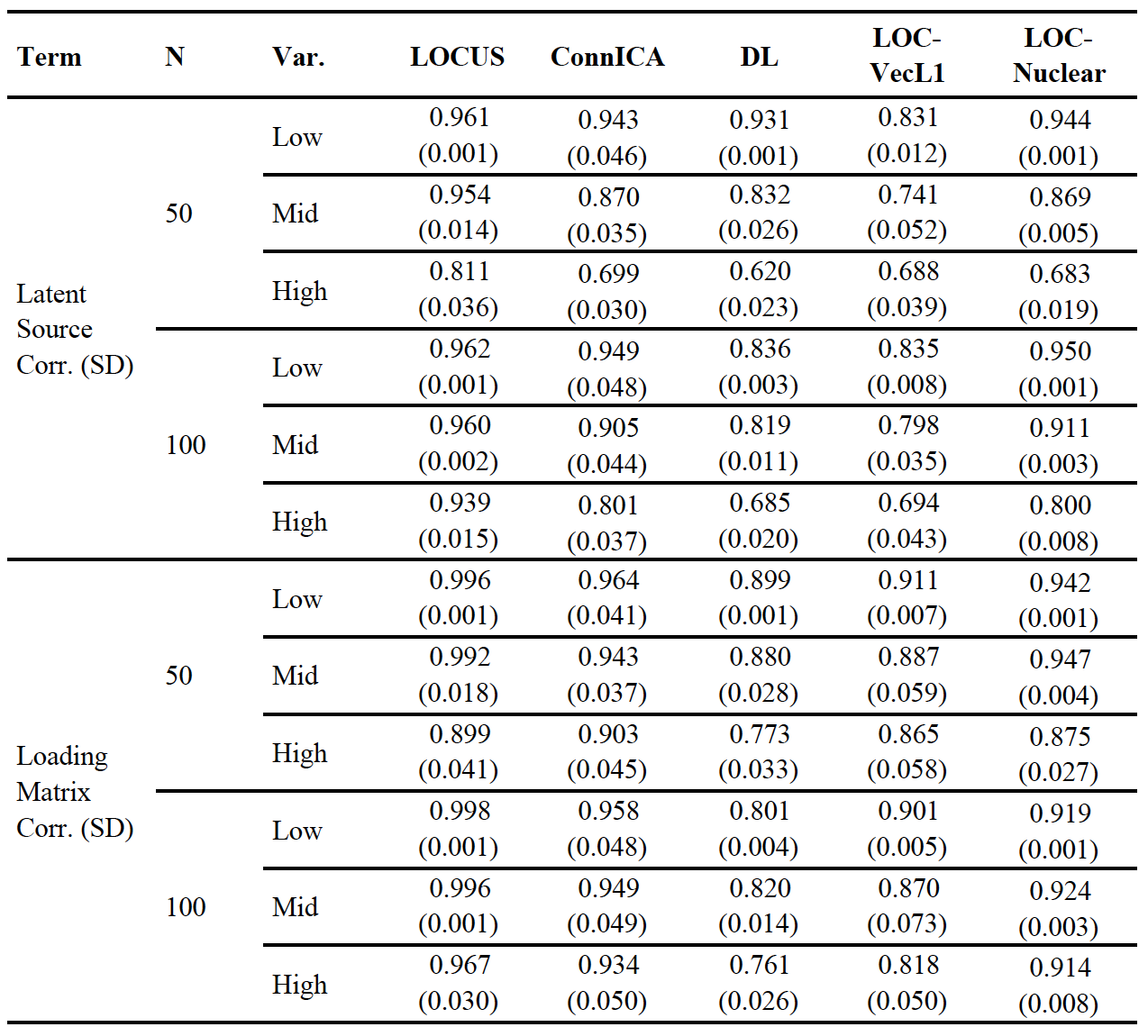}
\end{table}

\subsection{Evaluation Metrics}
\begin{figure}
\begin{center}
\small
\includegraphics[scale=0.5]{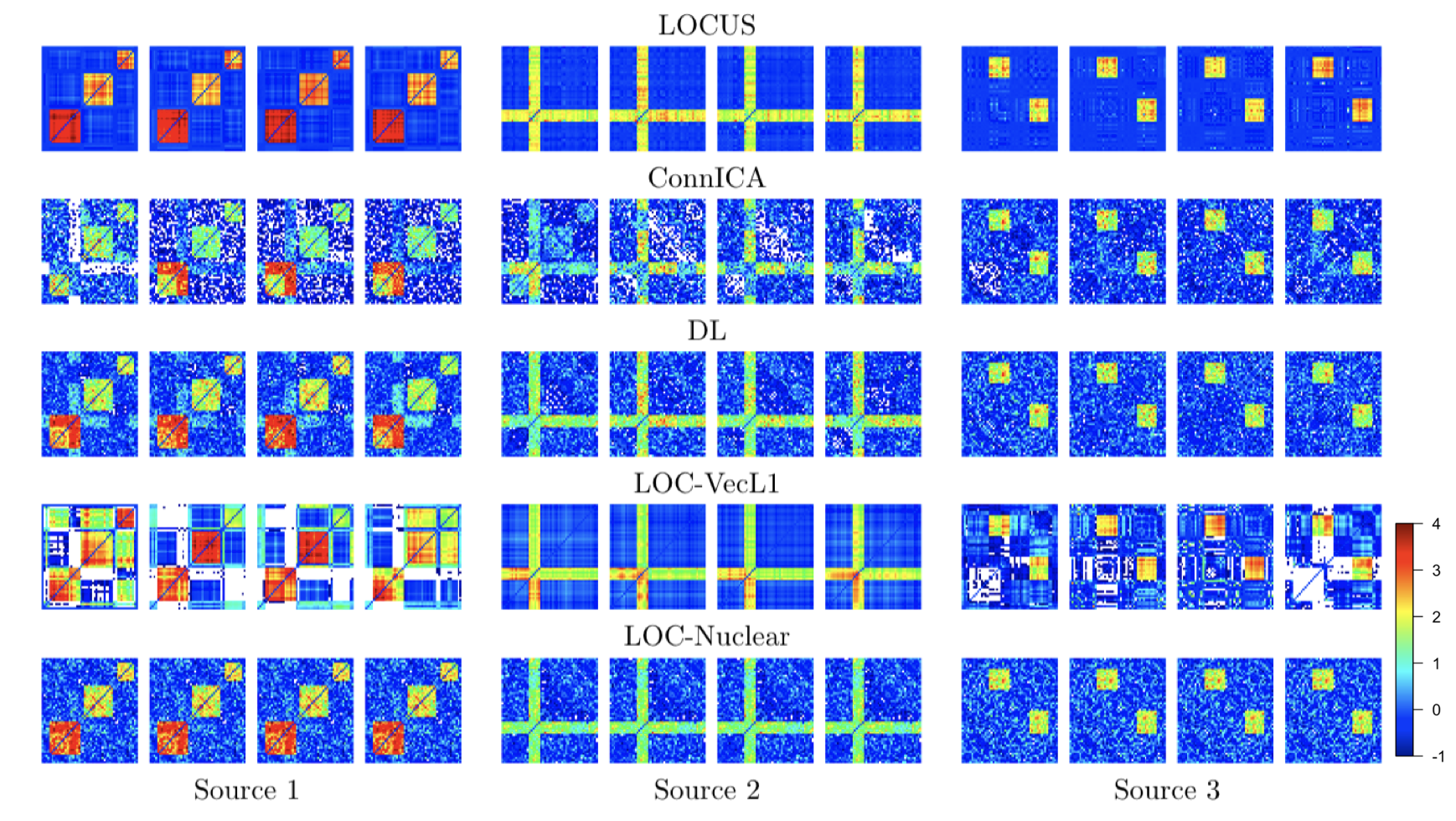}
\caption{\small{Estimated latent signals of 4 randomly selected simulation runs in Scenario I with high level variance across all methods. }}\label{fig:simuset1_pen_combine_highvar}
\end{center}
\end{figure}

\begin{figure}
\begin{center}
\small
\includegraphics[scale=0.5]{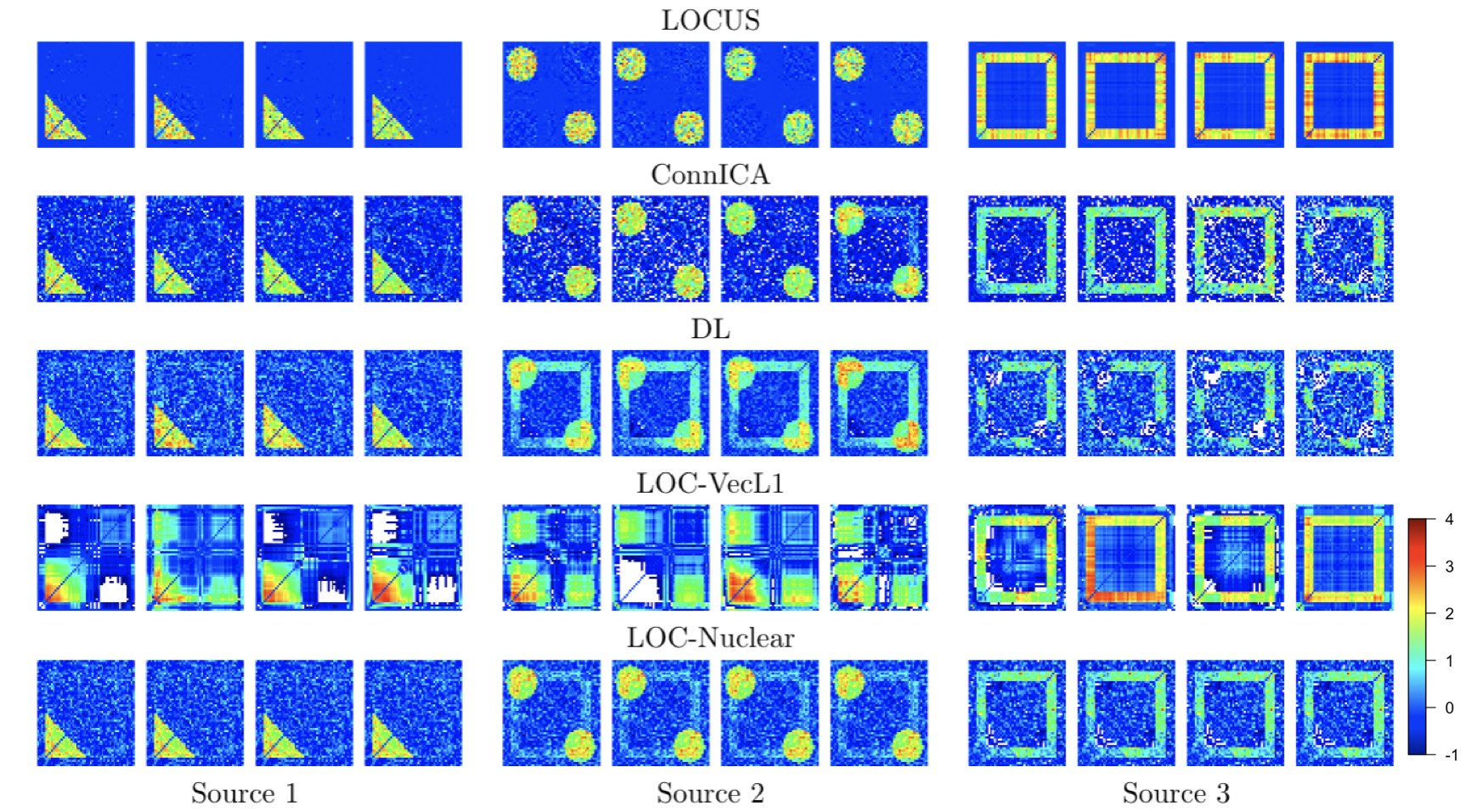}
\caption{\small{Estimated latent signals of 4 randomly selected simulation runs in Scenario II with high level variance across all methods. }}\label{fig:simuset2_pen_combine_highvar}
\end{center}
\end{figure}

Following previous work \citep{beckmann2005tensorial,guo2011general,wang2019hierarchical}, we evaluate the performance of each method
based on the correlations between the truth and the model-based estimates on the source signals and mixing coefficients. We further examine the standard deviation of the correlations across 100 simulation runs to evaluate the variability of the estimates. Results are summarized in Tables \ref{tbl:set1} and \ref{tbl:set2}.  To illustrate the accuracy of the methods in recovering the patterns in the latent sources, we randomly selected 4 simulation runs to display the estimated source signals for the high variance setting (Figures \ref{fig:simuset1_pen_combine_highvar} and \ref{fig:simuset2_pen_combine_highvar}). 

We evaluate the reproducibility of the latent sources extracted by each of the methods across the simulation runs using the reliability index in equation (\ref{eq:reproducibility}) where the replications correspond to the simulation runs with $B=100$, $\bm S_{\ell}$ is the true latent source, and $\hat{\bm S}^{(b)}_{\ell}$ is the latent source estimated from the $b$th simulation run that is matched with $\hat{\bm S_\ell}$. Figure \ref{fig:set1_reprod} displays the average reproducibility of the latent sources estimated via different methods under various simulation settings. 

\subsection{Simulation Results}

\subsubsection{LOCUS vs. the existing separation methods}
\begin{figure}
    \centering
    \hspace{2.5cm}  \includegraphics[scale = 0.45]{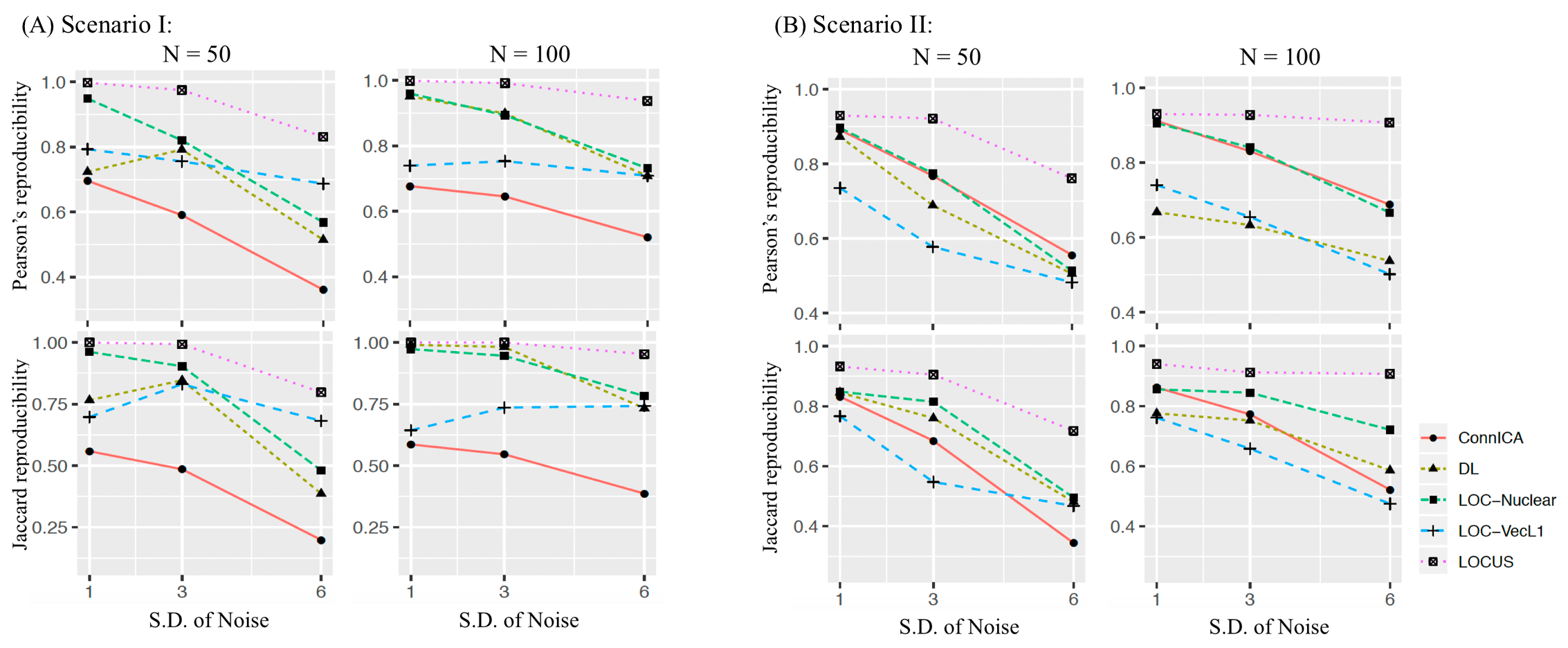}
    \caption{Reproducibility results on latent sources for both scenarios. The first row represents the average Pearson correlation between the true and estimated latent sources, while the second row represents jaccard index. }
    \label{fig:set1_reprod}
\end{figure}

Based on the results in Tables \ref{tbl:set1} and \ref{tbl:set2}, the proposed LOCUS method provides more accurate estimates for the latent sources and mixing coefficients or subject-specific loadings compared with the existing connICA and dictionary learning method. The correlation standard deviation of LOCUS is generally lower than that of connICA and dictionary learning, indicating the results from LOCUS have better stability. Figures \ref{fig:simuset1_pen_combine_highvar} and \ref{fig:simuset2_pen_combine_highvar} show that LOCUS produces more accurate results in recovering the underlying connectivity traits in both scenario I and II. In comparison, connICA generates considerably noisier estimates and shows cross-talking between some of its extracted traits. Dictionary learning has less accurate estimates than LOCUS in the activated region of the latent sources and also more false positive findings in the de-activated regions. Compared with LOCUS, the estimated latent sources of dictionary learning are noisy even with L1 penalty. This is because it does not model the sources with the low-rank structure as LOCUS does, which leads to considerably larger number of parameters to estimate and noisy results. Furthermore, latent sources extracted by the proposed LOCUS method consistently demonstrate higher reproducibility than those estimated by connICA and dictionary learning across all simulation settings (Figure \ref{fig:set1_reprod} shows). For example, in Scenario I, with sample size $N=100$ and high variance level, the correlation-based reliability index remains as high as 0.94 for LOCUS's estimates, while the reliability is only 0.52 and 0.71 for connICA and dictionary learning. 

\subsubsection{The novel sparsity regularization vs. the existing sparsity methods}

Results in Tables \ref{tbl:set1} and \ref{tbl:set2} show the proposed novel sparse regularization of LOCUS leads to more accurate estimates for the latent sources and subject-specific loadings compared with the existing vector-wise L1 regularization and the nuclear norm sparsity control. Figures \ref{fig:simuset1_pen_combine_highvar} and \ref{fig:simuset2_pen_combine_highvar} show that the proposed sparsity regularization produces more accurate and precise results in recovering the underlying connectivity traits in both scenario I and II. In comparison, the vector-wise sparsity leads to structured errors in estimating the latent sources and fails to detect the true patterns for some of the connectivity traits. The unsatisfactory performance is because the vector-wise sparsity aims to achieve element-wise sparsity in the vectors $\{\bm X_{\ell}\}$. In doing so, it often results in structured inaccuracy in estimating the latent sources which are inner products of the vectors. The nuclear norm sparsity method accurately recovers the activated regions in Scenario I where the latent sources have low-rank structure. However, its estimates are more noisy as compared with those of LOCUS and have more false positive signals in the non-activated regions. This is because the nuclear norm sparsity aims to achieve low-rankness in the estimated latent sources and does not necessarily lead to element-wise sparsity in the sources. For scenario II where the patterns are challenging to capture for the low-rankness structure, the performance of the nuclear norm sparsity is less satisfactory, showing cross-talking between the estimated sources. Finally, latent sources estimated based on proposed LOCUS sparsity regularization consistently demonstrate better reproducibility than those estimated with the vector-wise L1 penalty and nuclear norm penalty across all simulation settings (Figure \ref{fig:set1_reprod}).

\section{Discussion}\label{Discussion}

In this paper, we propose a novel signal separation framework designed for decomposing imaging-based brain connectivity matrices to reveal underlying connectivity traits. The proposed LOCUS method has several key innovations. Motivated by the observed characteristics in connectivity data, LOCUS uses a low-rank structure to significant reduce the number of parameters and improve accuracy and precision, leading to more efficient and reliable source separation for connectivity metrics. Moreover, we propose a novel angle-based sparsity regularization for the low-rank decomposition. This regularization is methodologically appealing by directly targeting the connectivity traits in its sparsity control, hence showing better performance than existing sparsity regularization methods. Furthermore, unlike many existing sparsity regularization which require numerical methods to solve, our new sparsity penalization lead to explicit analytic solutions to the optimization function in the estimation, which increases computational efficiency. Furthermore, our sparsity regularization is directly incorporated in the LOCUS optimization function. This is advantageous as compared with the two step approach in some other network models \citep{wang2019common} where a non-sparse estimator for the connectivity matrix is first obtained  and then
fed into an existing sparse method to obtain a sparse estimator. We show that the optimization function of LOCUS has the block multi-convex structure and propose a novel node-rotation algorithm for learning the LOCUS model. We conduct extensive simulation studies with data generated from various types of underlying source signals. The proposed LOCUS method demonstrates superior performance than the existing methods in both the simulation studies and the real data application.  The sparsity penalization term in the current paper is based on the L1 regularization. The proposed method and algorithm can be readily extended to alternative types of regularization such as L2, MCP and SCAD. Furthermore, the proposed LOCUS is applicable to various types of connectivity measures such as structural connectivity from DTI or functional connectivity measured by mutual information. Finally, the proposed angled-based sparsity regularization can be generally applied to tensor-decomposition methods that involve the low-rank structure, providing a useful new alternative to the existing sparsity penalization methods. In addition to the aforementioned advantages, the proposed LOCUS algorithm also has several appealing features. It generates highly reproducible results, is robust to different initiation values (Appendix D) and has great convergence performance (Appendix E). An R package for LOCUS will be released on publicly available websites such as The Comprehensive R Archive Network (CRAN) and Neuroconductor.

As noted in a paper, LOCUS is a probabilistic blind source separation method.  As with other noisy models such as the probabilistic ICA, the identifiability properties of such models are more elusive as compared with  noiseless models \citep{davies2004identifiability, eriksson2004identifiability, kagan1973characterization}. This is because the source signals can no longer be expressed as a direct function of the unmixing matrix and the data due to the presence of the noise term. Following previous work \citep{davies2004identifiability}, we can show that the mixing matrix $\bm A$ in the LOCUS model is identifiable up to a permutation and sign switching as in noiseless models. There is a degree of ambiguity in the full identifiability between the sources $\bm S$ and the noise, which is still an open research question for probabilistic blind source separation methods.

\begin{appendix}

\section{Proof of Lemma \ref{equ_prcess}:}
Denote $\mathcal{L}(\bm X_{\ell} \bm D_{\ell} \bm X_{\ell}')$ as $\bm s_{\ell}$, and given the orthogonality on $\widetilde{\bm A}$, we have
\begin{align}
    \sum_{i=1}^q\Big\| \widetilde{\bm y}_i - \sum_{\ell=1}^q \widetilde{a}_{i\ell} \mathcal{L}(\bm X_\ell \bm D_\ell \bm X_\ell')  \Big\|_2^2  &=  \sum_{i=1}^q\Big\| \widetilde{\bm y}_i - \sum_{\ell=1}^q \widetilde{a}_{i\ell} \bm s_\ell  \Big\|_2^2 = \| \widetilde{\bm Y} - \widetilde{\bm A} \bm S\|_F^2 = \| \widetilde{\bm A}(\widetilde{\bm A}'\widetilde{\bm Y} -  \bm S)\|_F^2 \nonumber \\
    &= \Big\|  [\widetilde{\bY}'\widetilde{\bm a}_{1},...,\widetilde{\bY}'\widetilde{\bm a}_{q}]' -  [\bm s_1,...,\bm s_q]'\Big\|_F^2 = \sum_{\ell = 1}^q \| \widetilde{\bY}'\widetilde{\bm a}_{\ell} -  \bm s_{\ell}  \|_2^2 \nonumber\\
     &= \sum_{\ell=1}^q \Big\| \widetilde{\bY}'\widetilde{\bm a}_{\ell} -  \mathcal{L}(\bm X_{\ell} \bm D_{\ell} \bm X_{\ell}') \Big \|_2^2, \nonumber
\end{align}
where $\widetilde{\bm a}_{\ell}$ is the ${\ell}$th column of $\widetilde{\bm A}$. This finished the proof of Lemma \ref{equ_prcess}. 

\section{Derivation of the update for $\bm x_{\ell}(v)$ in the node-rotation algorithm:} 

To see the linkage between (\ref{eq:locus_preprocess2}) and (\ref{update2})
, we first transfer (\ref{eq:locus_preprocess2}) into the following edge-wise form:
\begin{align} 
\Big\| &  \widetilde{\bm Y}'\widetilde{\bm a}_\ell - \mathcal{L}(\bm X_\ell\bm D_{\ell}\bm X_\ell') \Big\|_2^2 + \phi\sum_{u<v} |\bm x_\ell(u)'\bm D_\ell\bm x_\ell(v) |\nonumber \\
= & \sum_{u < v} \Big( \widetilde{\bm a}_\ell'\widetilde{\bm Y}(u,v) - \bm x_\ell(u)'\bm D_\ell \bm x_\ell(v) \Big)^2 + \phi\sum_{u<v} |\bm x_\ell(u)'\bm D_\ell\bm x_\ell(v) |,  \nonumber
\end{align}
which contains $V(V-1)/2$ terms. To update $\bm x_\ell(v)$ while conditioning on others, we only need the terms involving $\bm x_\ell(v)$ which contains $V-1$ terms:
\begin{align}
 f(\bm x_{\ell}(v)) &= \sum_{\substack{u=1 \\ u\ne v}}^V \Big( \widetilde{\bm a}_{\ell}'\widetilde{\bm Y}(u,v) - \bm x_{\ell}(u)'\bm D_{\ell} \bm x_{\ell}(v) \Big)^2 + \phi\sum_{\substack{u=1 \\ u\ne v}}^V  |\bm x_{\ell}(u)'\bm D_{\ell}\bm x_{\ell}(v) | \nonumber \\
 &= \Big\| \widetilde{\bm Y}_{\{v\}}'\widetilde{\bm a}_{\ell} -  \bm X_{\ell}(-v) \bm D_{\ell} \bm x_{\ell}(v) \Big\|_2^2 + \phi \sum_{\substack{u=1 \\ u\ne v}}^V | \bm x_{\ell}(u)' \hat{\bm D_{\ell}} \bm x_{\ell}(v)  |,\nonumber 
\end{align}
which is same as (\ref{update2}). 

\section{Proof of Proposition \ref{Theo:biconvex}. Block Multi-Convexity:}

It is straightforward to show the following proof of block multi-convexity applies to both the LOCUS model (\ref{eq:LocusICAgroup}) on the original data as well as the model (\ref{eq:LocusICAgroup_preproc}) on the preprocessed data. We will use the notations from the original data model in the following. 

Let $f(\cdot)$ be the objective function in (\ref{eq:locus_preprocess}). We define the parameter partition of $\mathcal{P} = \{ \bm x_1(1),..,\bm x_1(V),$ $..,\bm x_q(V), \bm d_1,..,\bm d_q,\bm A \}$, where $\bm x_{\ell}(v) (v=1,\ldots,V)$ is the $v$th element of $\bm X_\ell\,(\ell=1,\ldots,q)$ and $\bm d_\ell=\textrm{diag}(\bm D_\ell)$. We show that the function $f$ is convex with respect to each individual argument in $\mathcal{P}$ while holding the others fixed. 

First, we show the convexity of $f$ w.r.t. $\bm x_{\ell}(v)$ given the other terms. We rewrite the function $f$ as follows,
\begin{align} 
\label{eq:multiconvex_proof_f}
f &= \sum_i \sum_{u<v} \Big(\bm y_i(u,v)- \sum_{\ell=1}^q a_{i\ell}\bm x_{\ell}(u)'\bm D_{\ell}\bm x_{\ell}(v)  \Big)^2 + \phi \sum_{{\ell}=1}^q\sum_{u<v}|\bm x_{\ell}(u)'\bm D_{\ell}\bm x_{\ell}(v)| \nonumber \\
&= \sum_{i} \sum_{\substack{u=1 \\ u\ne v}}^V \Big(\bm y_i(u,v)- a_{i{\ell}}\bm x_{\ell}(u)'\bm D_{\ell}\bm x_{\ell}(v) -\sum_{h\ne \ell}a_{ih}\bm x_h(u)'\bm D_h\bm x_h(v)    \Big)^2 + \phi \sum_{\substack{u=1 \\ u\ne v}}^V|\bm x_{\ell}(u)'\bm D_{\ell}\bm x_{\ell}(v)| + c \nonumber \\
&= \sum_{i} \Big( \sum_{\substack{u=1 \\ u\ne v}}^V \bm y_{i\{v\}}^{(0)} - a_{i\ell} \bm X_{\ell}(-v) \bm D_{\ell}\bm x_{\ell}(v)\Big)^2  + \phi \|\bm X_{\ell}(-v)\bm D_{\ell}\bm x_{\ell}(v)\|_1 + c, 
\end{align}
where $\bm y_i(u,v)$ is the element in $\bm y_i$ that corresponds to the edge connecting node $u$ and $v$,
$\bm y_{i\{v\}}^{(0)} = \bm y_i(u,v)-\sum_{h\ne \ell}a_{ih}\bm x_h(u)'\bm D_h\bm x_h(v)$ and $c$ is a constant which doesn't involve $\bm x_{\ell}(v)$.

From (\ref{eq:multiconvex_proof_f}), we can derive the Hessian matrix of $\bm x_{\ell}(v)$ to be $2\sum_{i=1}^N  a_{i\ell}^2 \bm D_{\ell} \bm X_{\ell}(-v)'\bm X_{\ell}(-v)\bm D_{\ell}$, which is positive semi-definite. This proves the convexity of $f$ w.r.t. $\bm x_\ell(v)$ while fixing the other terms. 

Next, to show that $f$ is convex w.r.t. $\bm d_\ell$ while fixing the other terms, we note that $\bm x_\ell(u)'\bm D_\ell \bm x_\ell(v) = (\bm x_\ell(u) \circ \bm x_\ell(v))'\bm d_\ell$, where $\circ$ denotes Hadamard product. We can readily derive the Hessian matrix for $\bm d_\ell$ which can also be shown as positive semi-definite. Lastly, for $\bm A$, when fixing the other terms, the function $f$ in terms of $\bm A$ is the least square problem and hence the convexity of $f$ w.r.t. $\bm A$ readily follows.

\section{Additional Results for the PNC Study:}
\begin{figure}
\begin{center}
\small
\includegraphics[scale=0.3]{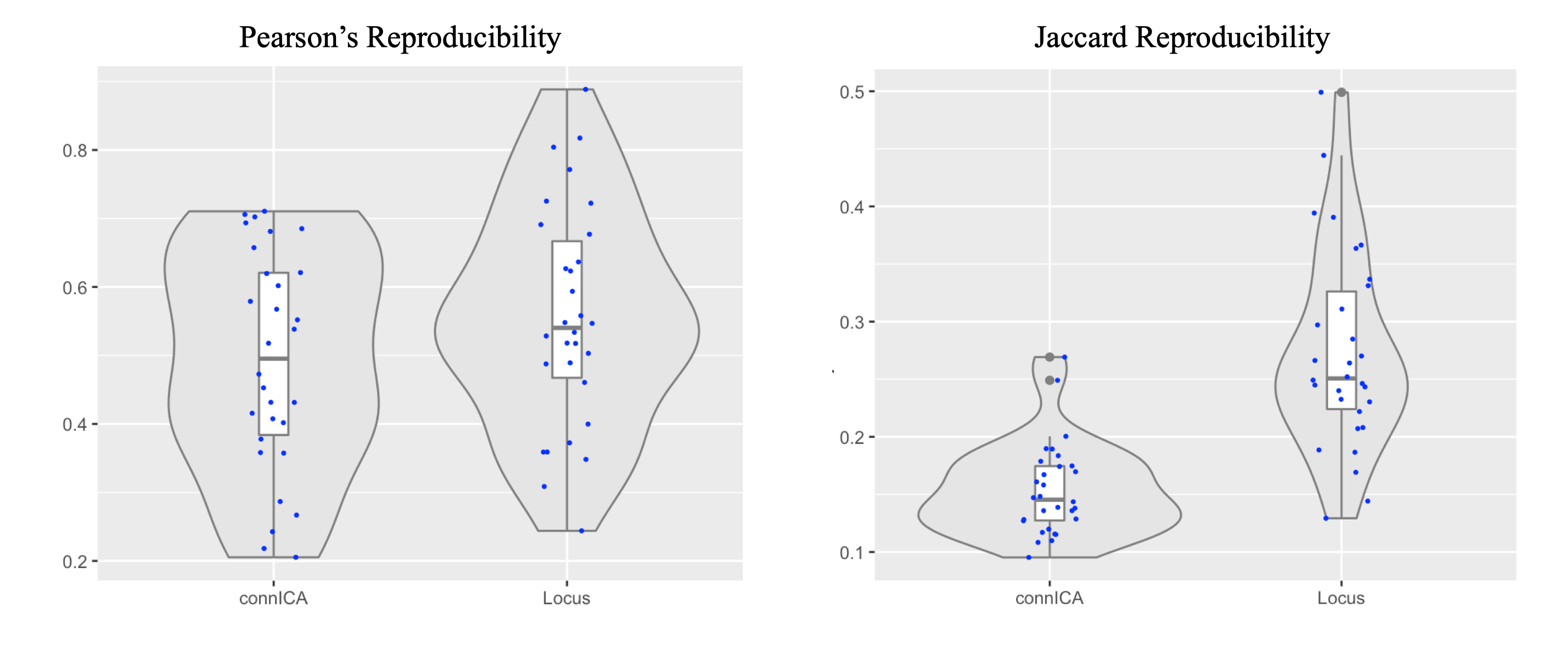}
\caption{\small{Reproducibility analysis for all 30 latent sources from LOCUS and connICA.}}\label{fig:reprod_all}
\end{center}
\end{figure}

\begin{figure}
\begin{center}
\small
\includegraphics[scale=0.5]{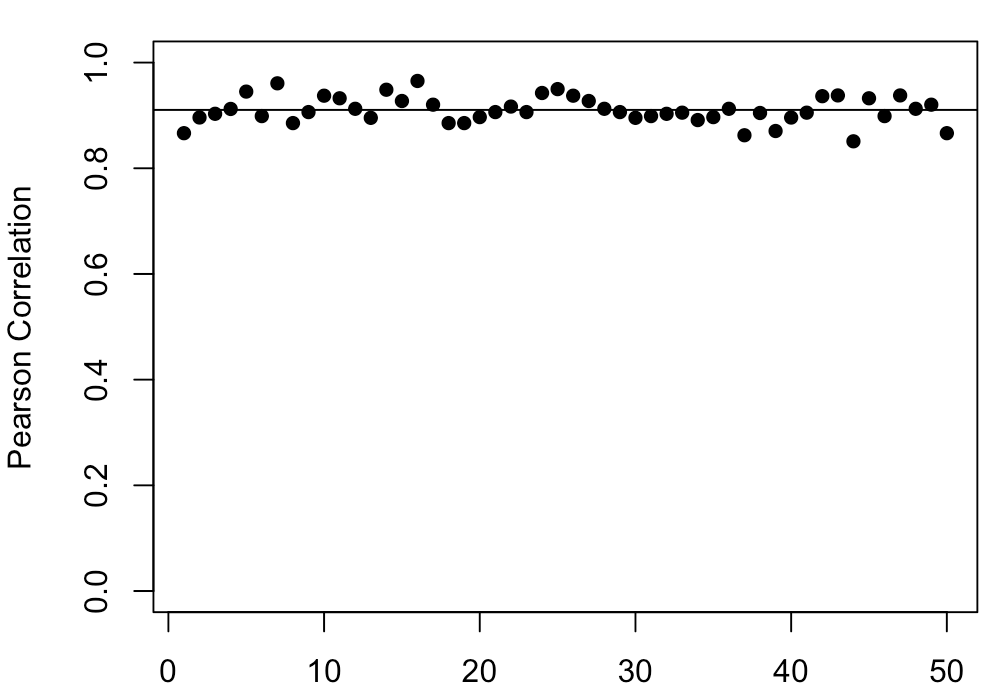}
\caption{\small{Reproducibility of the LOCUS extracted latent sources in PNC Study with respect to initialization. Values presented are the cross-source average correlation between the original LOCUS results reported in the paper and the LOCUS results from each of the 50 different initializations.}}\label{fig:50_initialization}
\end{center}
\end{figure}

This section contains additional results from the PNC study.  Figure \ref{fig:reprod_all} shows the distribution of Pearson and Jaccard based reproducibility measures for all the 30 latent sources extracted by LOCUS and connICA. As shown in the figure, LOCUS generally has higher reliability than connICA. The better reproducibility of LOCUS is more clearly demonstrated with the Jaccard index which reflects the reproducibility in identifying the significant edges in each connectivity trait. We also evaluate the reproducibility of the results from LOCUS in the PNC Study with respect to initialization. We run LOCUS on the PNC dataset using 50 different initializations, and evaluate the correlations between the LOCUS extracted source signals originally reported in the paper with the matched source signals extracted by LOCUS using each of the 50 initializations.  Figure \ref{fig:50_initialization} presents the cross-source average correlation between the original results and the results from each of the 50 different initializations. LOCUS results are fairly robust to different initialization with an average correlation of 0.91 across the 50 runs. 

\section{The Convergence of the LOCUS Algorithm}

We conducted simulation and real data studies to evaluate the convergence rate of the LOCUS algorithm across various scenarios. First, in the simulation study, we considered the same simulation settings as in Section \ref{sec3} and evaluated the convergence rate across 100 simulation runs per setting with the criterion  $\|\hat{\bm\Theta}_{t+1} - \hat{\bm\Theta}_t \|_F / \hat{\|\bm\Theta}_t\|_F < 0.001$. %where $\bm\Theta = (\bm S, \widetilde{\bm A})$. 
The LOCUS algorithm converged in all simulation runs within $100$ iterations across all the settings. Furthermore, we check the convergence rate of LOCUS in the PNC study using 50 different initialization. The LOCUS algorithm converged within $200$ iterations across all the 50 initialization.

\section{Additional Simulation Studies with  varying signal intensity for the latent sources}
\begin{figure}
    \begin{center}
    \small
    \includegraphics[scale=0.4]{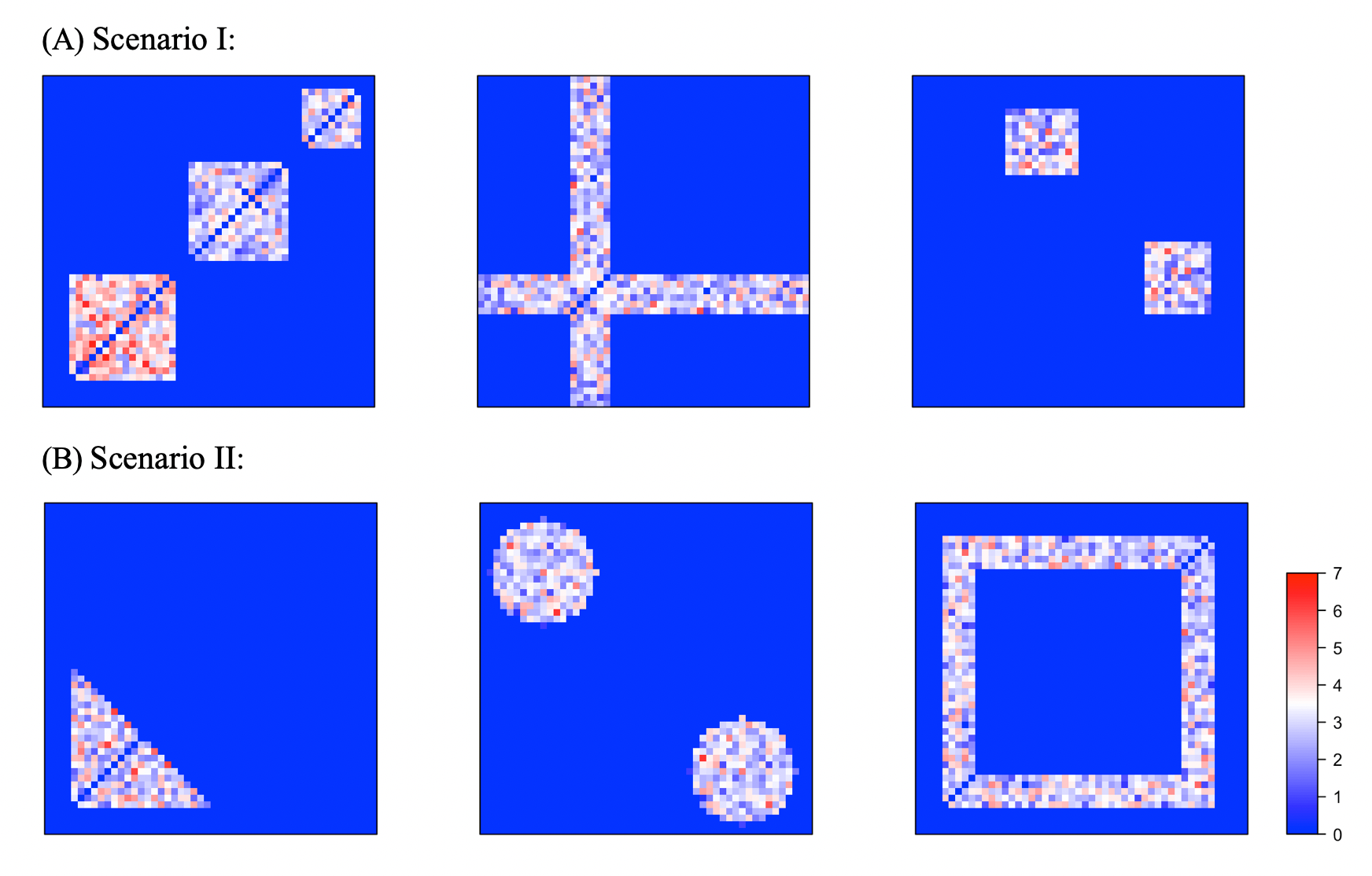}
    \caption{\small{True underlying source signals for two simulation scenarios in the additional simulation studies with varying signal intensities.}}\label{fig:noisysimusetting}
    \end{center}
\end{figure}

This section contains additional simulation results where there are variations in the signal intensity across edges in the latent sources $\bm S$ (Figure \ref{fig:noisysimusetting}). Table \ref{tbl:noisy_S} presents the results from LOCUS and connICA. Generally, estimates from LOCUS show better accuracy and smaller variability than those from connICA, which is a similar finding as from the original simulation studies in the paper.

\begin{table}
	\caption{Simulation results for comparing LOCUS with connICA with 100 simulation runs for Scenario I and II in the additional simulation studies with varying signal intensities. Values presented are mean and standard deviation of correlations between the true and estimated latent sources and loading/mixing matrices.}
	\label{tbl:noisy_S}
	\centering
		\includegraphics[scale=0.65]{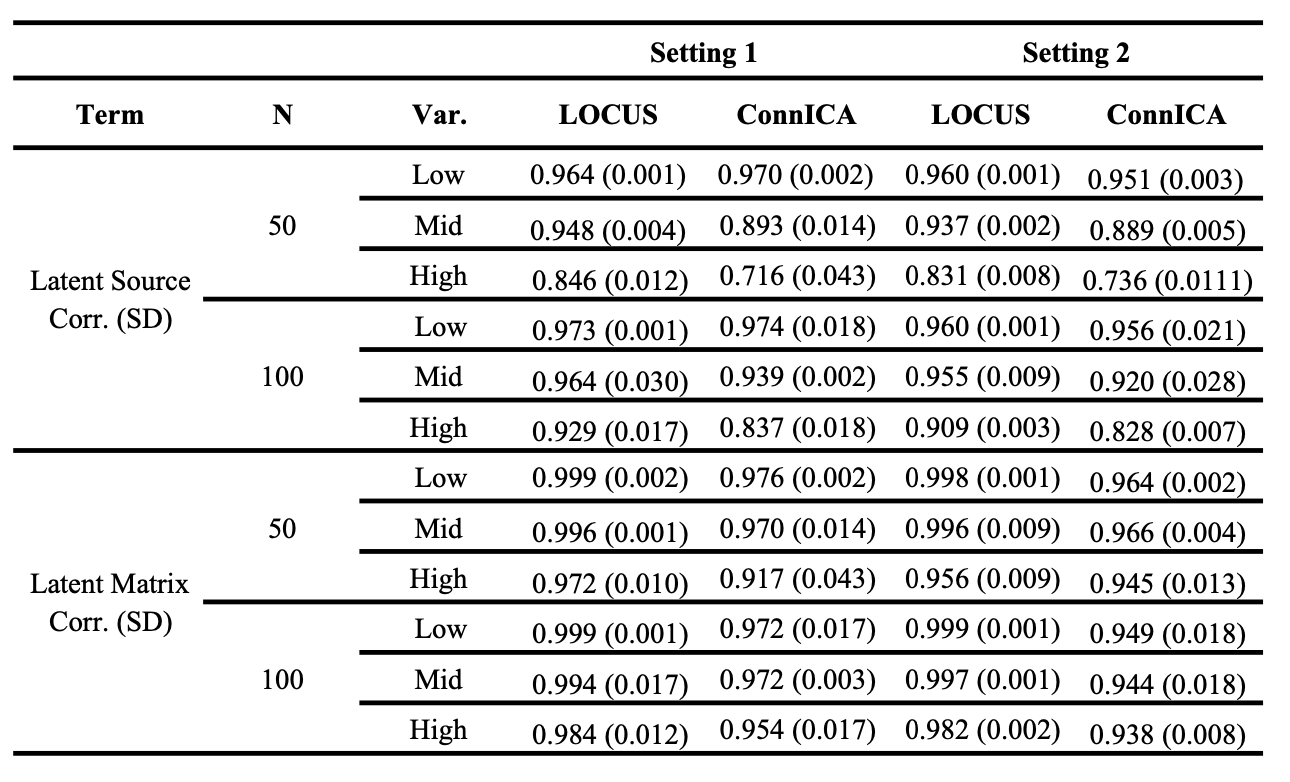}
\end{table}

\end{appendix}
\section*{Acknowledgements}

Research reported in this publication was supported by the National Institute of Mental Health of the National Institutes of Health under Award Number R01MH105561 and R01MH118771. The content is solely the responsibility of the authors and does not necessarily represent the official views of the National Institutes of Health. Philadelphia Neurodevelopmental Cohort: Support for the collection of the data sets was provided by grant RC2MH089983 awarded to Raquel Gur and RC2MH089924 awarded to Hakon Hakorson. All subjects were recruited through the Center for Applied Genomics at
The Children's Hospital in Philadelphia.

\bibliographystyle{imsart-nameyear}
\bibliography{locus_aoas_final_print_ver1}

\end{document}